\documentclass[a4paper]{article}
\usepackage{amsmath,amssymb}
\usepackage{stmaryrd}
\usepackage{amsthm,bm}
\usepackage{amsfonts,mathtools}
\usepackage[cp1251]{inputenc}
\usepackage{graphicx,subfigure}
\usepackage{epsfig,color,pifont,cite,enumerate}
\usepackage{setspace}
\graphicspath{{img/}}

\usepackage{geometry}
\DeclareMathAlphabet{\mathpzc}{OT1}{pzc}{m}{it}
\newcommand{\V}{\mathpzc{V}}
\newcommand{\ddd}{\mathpzc{d}}
\newcommand{\hn}{\bldh_{\text{tan}}}
\newcommand{\mk}{}

\newcommand{\grad}{\nabla\!\!\!\!\nabla}
\newcommand{\fff}[2]{\bm{II}\!\left[ #1; #2 \right]}
\newcommand{\bldh}{\boldsymbol{h}}
\newcommand{\st}{\,\dot{\mathfrak{s}}}

\usepackage[mathscr]{eucal}
\renewcommand{\phi}{\varphi}
\renewcommand{\epsilon}{\varepsilon}
\renewcommand{\kappa}{\varkappa}
\newcommand{\br}{\mathbb{R}}
 \newcommand{\bldr}{\boldsymbol{r}}
 \newcommand{\proj}{\text{\bf Pr}}

\newtheorem{theorem}{Theorem}[section]

\newtheorem{assumption}{Assumption}[section]
\newtheorem{lemma}{Lemma}[section]

\newcommand{\oms}{\text{\scriptsize $\mathcal{O}$}}
\newcommand{\bldv}{\boldsymbol{v}}
\newcommand{\blda}{\boldsymbol{a}}
\newcommand{\bldp}{\boldsymbol{p}}
\newcommand{\bldq}{\boldsymbol{q}}
\newcommand{\blde}{\boldsymbol{e}}
\newcommand{\btau}{\boldsymbol{\tau}}
\newcommand{\bn}{\boldsymbol{n}}

\newcommand{\bro}{\boldsymbol{\varrho}}

\newcommand{\sgn}{\text{\bf sgn}\,}

\newcommand{\bldu}{\boldsymbol{u}}

\newcommand{\so}{\text{\scriptsize $\mathcal{O}$}}
\newcommand{\spr}[2]{\left\langle #1; #2 \right\rangle}

\newcommand{\ov}[1]{\overline{#1}}

\newcommand{\dd}[1]{#1^{\prime\prime}}

\newcommand{\pf}{{\bf Proof:}\;}
\newcommand{\epf}{$\qquad \Box$}

\newcommand{\pp}{{\prime\prime}}

\title{Robotic Following of Flexible Extended Objects: Relevant Technical Facts on the Kinematics of a Moving Continuum}
\date{}
\author{Alexey S. Matveev\footnotemark[2] and Valentin V. Magerkin\footnotemark[2]}
\begin{document}
\UseRawInputEncoding
\maketitle
\renewcommand{\thefootnote}{\fnsymbol{footnote}}
\footnotetext[2]{Department of Mathematics and Mechanics, Saint Petersburg State University, St. Petersburg, Russia}
\section{Introduction}
This paper is motivated by general issues concerned with autonomous navigation, coordination, and motion control of formations of mobile robots in 2D and 3D.
Over the past decades formation control has reached a considerable level of maturity \cite{ReCa10,OhPaAh15}.
Meanwhile, the focus of interest was noticeably shifted towards the situation where coordination should be achieved by a formation that includes not only mobile agents jointly pursuing a common objective but also those apathetic or even hostile to it, which act non-cooperatively and on their own.
\par
Among numerous examples of such situations, there is the problem of autonomously driving planar robots into a formation encircling an independent targeted object, see, e.g., \cite{SiGh07,MBF06,KiSu07,HiNa09,KiHaHo10tr,AhFiAnd11,KoHo11,Yamaguchi03,KoShPo13,TsHaSaKi10,CeMaGaGi08,GYL10,LYL10,ZaMaHoSa14} and literature therein.
A motivation of this problem comes from many sources, including transportation of large objects, rescue operations, exploration and surveillance missions, minimization of security risks and upgrading situational awareness before coming to a closer contact, deployment of mobile sensor/actuator networks \cite{SaChXiJaMa15}, escorting and patrolling missions, troop hunting, to name just a few. This diversity gives rise to many different incarnations of the targeted object.
This may be a single target or a group of them; in some cases, the targets may be treated as point-wise objects, whereas they are extended bodies in other situations. Some scenarios deal with steady bodies, whereas other ones involve moving objects; their motions may be rigid in some cases and include changes in the shape and size (deformations) in other cases.
Various combinations of these are also feasible.
\par
In many missions of the considered type, the robots should autonomously detect and localize the targeted object, then to get near it and finally arrive at positions that are tactically advantageous for attaining the posed objective. For planar work-spaces, the locus of such positions is often an object-dependent curve, moving and deforming in general. With this in mind, the navigation objective includes and considerably comes to arriving at this curve and to subsequently tracking it. In typical multi-robot scenarios, the robots should also
autonomously achieve and then maintain an effective self-distribution over this curve, e.g., an even one, so that the robotic team surrounds the targeted object more or less uniformly. For example, if the target is single and point-wise and the advantageous position is delineated as that at a certain specific distance from the target, the discussed dynamic curve is the circle centered at the target whose radius equals that distance. If the targeted object is an extended planar body and a profitable position is still at a given distance from it, the curve is constituted by the points equidistant from this object. In the case of multiple objects, the ``distance from it'' may be replaced by the ``mean distance from them'' or the ``distance to the nearest of them'' in these examples.
\par
In the scenarios sketched out above, a certain moving curve comes to the fore in stating the control objective. By its own right,
this beneficially imparts a certain degree of uniformity to the problem setup and analysis by permitting one to neglect many details of diverse ``true'' targeted objects, which may  substantially vary from case to case, during the basic part of design and verification of a navigation algorithm. Anyhow, kinematical and other characteristics of the above curve inevitably play a major role in mathematically rigorous justification of the algorithm. From a general perspective, this curve is a dynamic continuum undergoing a general motion, and there is little to be added to this due to the diversity of the scenarios. Up to now, such continua were mostly studies within the framework of the continuum mechanics \cite{Spencer04,AltOch20}. However, this discipline is mostly focused on the mechanical behavior of materials modeled as a continuous mass, and is not much concerned with navigation of robots relative to continua. As a result, studies on the last issue systematically face unavailability of presumably basic formulas, whose derivation is somewhat lengthy so that it seems out of place when putting in research articles on robotics. This paper is aimed at filling this gap with respect to kinematics of moving planar curves.
\par
The second part of this paper pursues a similar objective with respect to kinematics of moving and deforming level sets of dynamic scalar environmental fields.
This issue is inherent in dealing with the problem of robotic detection, localization, and exploration of such sets. This problem is also a subject of an extensive research; see e.g., \cite{QBGG05,AKP06,GHH04,WSS08,WTASZ05,CKBMLM06,HJMNTBM05,BertKeMar04,ClaFie07,JAHB09,PeDuJoPo12,MaBe03,SuBuMa08,SuLiHe12,KrNi16} and the literature therein. A typical relevant mission is to find and arrive at an a priori unknown level set where an environmental field assumes a specific threshold value, and then to cover the entirety of this set. This results in exhibiting and gaining control over the border of the region with greater field values, which is commonly the major focus of interest.
Examples include tracking of oil or chemical spills, or other contaminants \cite{ClaFie07tr}, detection and monitoring of harmful algae blooms \cite{PeDuJoPo12}, tracking zones of turbulence, contaminant clouds \cite{WTASZ05}, or high radioactivity level, exploration of sea salinity and temperature or hazardous weather conditions, to name just a few.
Up to now, the research on robotic tracking of environmental level sets mostly dealt with 2D work-spaces.
Meanwhile, expansion of underwater, flying, and space robots motivates strong interest to autonomous navigation with using all three dimensions and to cases where the problem cannot be reduced to a 2D setting. Handling these issues inevitably addresses kinematics of isosurfaces in 3D, which is the second topic of this paper.
\par
The body of the paper is organized as follows. Section~\ref{sec1} is motivated by the problem of robotic following of planar curves
and offers a number of technical and relevant to the issue facts about their kinematics.
Section~\ref{sec.3d} addresses kinematics of isosurfaces and is inspired by the problem
of their sweeping coverage in the case of a  time-varying environmental field in 3D.
Section~\ref{sec.drhs} presents a technical fact that extends Theorem~4.1 in Chapter III \cite{Hart82} on ODE with discontinuous right-hand sides.

\section{Following an Unpredictably Moving Speedy Planar Curve}
\label{sec1}
\subsection{Motivation and Assumptions}
The interest in the developments of this section largely arises from the following scenario.
Several planar robots travel in the plane and are driven by the acceleration vectors. The plane also hosts a moving and deforming Jordan curve $\Gamma=\Gamma(t)$, which is unknown, unpredictable and maybe, speedy. The robots should reach this curve and then trace it in a common given direction. An effective self-distribution over $\Gamma$ should be achieved, with the even one being an ideal option.
\par
Some examples of pertinent missions are as follows.
\par
{\bf 1.} There is a moving 2D continuum $D(t) \subset \br^2$ of arbitrary and time-varying shape. This covers scenarios with reconfigurable rigid bodies, forbidden zones between vehicles moving in a platoon, flexible obstacles, like fishing nets or schools of fish, virtual obstacles, like on-line estimated areas of threats or areas contaminated with chemicals or corrupted with a high turbulence.
The robots should advance at a given distance $d_0>0$ to $D(t)$, then maintain it, circumnavigate $D(t)$ in a common direction, and form a dynamic envelope of $D(t)$ via uniformly, more or less, surrounding $D(t)$. In this case, $\Gamma(t)$ is the locus of points at a distance of $d_0$ from $D(t)$.
\par
{\bf 2.} There are multiple speedy and unknowingly moving pointwise targets $\bldp_j(t)$. It is needed to drive the root mean square distance $d_{\text{mean}}$ from every robot to the targets to a given value $d_0$, to effectively distribute the robots over the locus  $\Gamma(t)$ of points with $d_{\text{mean}}=d_0$, and to subsequently follow the targets with maintaining this value and distribution.
\par
{\bf 3.} In the case {\bf 2}, the targets should be fully enclosed and tightly circumnavigated:
a given distance $d_0$ to the currently nearest target should be reached and maintained by every robot, while all targets are to be inside its path.
In this case, $\Gamma(t)$ is composed of arcs of $d_0$-circles $\{\bldr \in \br^2: \min_i \|\bldr - \bldp_i(t)\| = d_0\}$ centered at the targets, where $\|\cdot\|$ is the standard Euclidean norm. The targets are assumed not to spread too far apart from each other so that such arcs can be concatenated to form a non self-intersecting loop $\Gamma(t)$ encircling all targets. Since this curve is typically non-smooth and the robots can trace only smooth paths with nonzero speeds, $\Gamma(t)$ should be approximated by a smooth curve to make the mission feasible; see \cite{MaSeSa17} for details.
\par
{\bf 4.} In the missions 2 or 3, the targets are not point-wise but are moving 2D continua $D_j(t) \subset \br^2$ of arbitrary and time-varying shapes.
In this example, the mean squared distance is defined as $\sum_j d_j(t)^2$ and $d_j(t)$ is the distance from the robot to the closest point of $D_j(t)$.
\par
{\bf 5.} In the mission 2, the targets are moving and deforming $2D$ bodies $D_j(t) \subset \br^2$, and the mean squared distance
to the points of all (say $N$) targeted bodies is considered; this distance is given by
\begin{gather*}
\int_{D_1(t) \cup \ldots \cup D_N(t)} \|\bldr - \bro\|^2 \; d \bro .
\end{gather*}
\par
{\bf 6.} The plane hosts an a priori unknown scalar time-varying field, which is represented by a function $F(\bldr,t) \in \br$ of spatial location $\bldr \in \br^2$ and time $t \in \br$. For example, this field may describe the concentration of a contaminant, the level of radiation, the strength of an acoustic or electromagnetic signal, the temperature, etc. The robots should locate and advance to the isoline (level set) $\Gamma(t) := \{\bldr: F(\bldr,t) = f_0\}$ where the field assumes a pre-specified value $f_0$. After reaching the isoline, the robots should track it and evenly distribute themselves over its length for the purposes of displaying, monitoring, or processing this curve.
\par
{\bf 7.} Mission 6 is troubled by the presence of moving obstacles $O_1(t), \ldots, O_N(t)$ in the scene, which should be bypassed with respecting a desired safety margin $d_{\text{safe}}$. For the sake of brevity, we do not specifically discuss construction of the targeted curve $\Gamma(t)$ and only remark that it is reasonable to compose $\Gamma(t)$ via concatenation of ``free-space'' pieces of the isoline with fragments of the $d_{\text{safe}}$-equidistant curves of the obstacles; see \cite{MaNi21} for details. Since the resultant curve is typically non-smooth, it should be approximated by a smooth curve due to the reasons discussed in 3.
\par
In this paper, we neglect the origins and genesis of the moving curve $\Gamma(t)$ and treat it as a self-sufficient component of the scene, which defines the navigation objective of the robots. To describe this curve, we use the Lagrangian approach \cite{Spencer04} and so introduce a {\it reference configuration} $\Gamma_{\text{ref}}$ and a time-varying {\it configuration map} $\Phi(\cdot,t)$ that transforms $\Gamma_{\text{ref}}$ into the {\it current configuration} $\Gamma(t) = \Phi[\Gamma_{\text{ref}},t]$. We limit the motion of the curve $\Gamma(t)$ by only a few and minimal conventions, typical for the general continuum mechanics \cite{Spencer04}.
\begin{assumption}
\label{ass.obs}
The set $\Gamma_{\text{\rm ref}} \subset \br^2$ is a $C^3$-smooth Jordan curve; the map
$\Phi(\cdot,t)$ is defined on an open connected vicinity $O_\ast$ of $\Gamma_{\text{\rm ref}}$, is $C^3$-smooth and one-to-one, its Jacobian matrix $\Phi_{q}^\prime(\cdot,t)$
is everywhere invertible.
\end{assumption}
\par
The velocity $V(\bldq,t)$ and acceleration $A(\bldq,t)$ of the moving point $\bldq=\bldq(t) \in \Gamma(t)$ are given by
\begin{equation}
\label{def.velacc}
V(\bldq,t) := \frac{\partial \Phi}{\partial t}[\bldq_\ast,t], \qquad A(\bldq,t) := \frac{\partial^2 \Phi}{\partial t^2}[\bldq_\ast,t],
\end{equation}
where $\bldq_\ast \in \Gamma_{\text{ref}}$ is the ``seed'' of the ``particle'' $\bldq(t) = \Phi(\bldq_\ast,t)$.
\par
The next claim is usually fulfilled in the real world.
\begin{assumption}
\label{ass.obs1}
The configuration map $\Phi(\cdot)$, its first, second, and third derivatives, and the inverse to the spatial Jacobian matrix $\Phi_{q}^\prime(\cdot)$ are bounded on the domain of definition $O_\ast \times [0, \infty)$.
\end{assumption}
\par
We model robot $i$ as a double integrator
\begin{equation}
\label{1}
\ddot{\bldr}_i = \blda_i, \quad \bldr_i(0) = \bldr_i^0, \quad \bldv_i(0) = \bldv_i^0. \end{equation}
Here $\bldr_i$ is the position of the $i$th robot, $\bldv_i$ is its velocity, whereas its acceleration $\blda_i$ is the control input.
\par
Let $d_i$ stand for the signed distance from robot $i$ to the curve $\Gamma(t)$, which distance is defined to be positive/negative outside/inside the area bounded by the curve.
In terms of this distance, the task of reaching the curve takes the form $d_i \to 0$.
To regulate the output $d_i$ to the desired value $0$, {\it local controllability} of the output is classically required. At the least, this trait means that respective controls can cause keeping the distance to $\Gamma$ constant, converging to $\Gamma$, and diverging from $\Gamma$. Since the relative degree of the output $d_i$ is $2$, this means that whenever $\dot{d}_i=0$, the sign of $\ddot{d}_i$ can be arbitrarily manipulated by means of feasible accelerations. We assume this only at the maximal speed and everywhere in the {\it operational zone}.
For the sake of convenience, this zone is delineated in terms of the distance $d$ to $\Gamma$:
\begin{equation}
\label{zop}
Z_{\text{op}} := \big\{ (\bldr,t) : d_- < d < d_+ \big\}, \qquad \text{where} \quad d_- <0 < d_+
\end{equation}
are given. By \cite[Lem.~3.1]{MTS11}, the capacity to maintain the distance to $\Gamma$ implies the following properties (up to a minor enhancement of the second of them).
\begin{assumption}
\label{ass.focal}
At any time $t$ and for any point $\bro \in \Gamma(t)$ of the curve, the following inequality holds:
\begin{equation}
\label{ineq.focus}
0 < 1+\varkappa(\bro,t) d_{-\sgn \varkappa}.
\end{equation}
At any time $t$, the distance from any point $\bldr$ of the operational zone (i.e., such that $(\bldr,t) \in Z_{\text{\rm op}}$) to the curve $\Gamma(t)$ is furnished by a single point of this curve.
\end{assumption}

\subsection{Some geometric and kinematic formulas concerned with moving planar curves}
We use the following notations:
\begin{itemize}
\item $\spr{\cdot}{\cdot}$ and $\|\cdot\|$, standard Euclidean inner product and norm, respectively;
\setlength{\itemsep}{-0.3\baselineskip}
\item $\pi[\bldr,t]$, {\it projection} of point $\bldr$ onto the curve $\Gamma(t)$, i.e., the point of $\Gamma(t)$ nearest to $\bldr$;
\item $\bro_i(t) := \pi[\bldr_i(t),t]$, projection of the current location of robot $i$ onto the curve $\Gamma(t)$;
\item $\mathpzc{d}(\bldr,t)$, unsigned distance $\min_{\bro \in \Gamma(t)} \|\bldr - \bro\|$ from point $\bldr$ to the curve $\Gamma(t)$;
\item $d(\bldr,t)$, signed distance from point $\bldr$ to the curve $\Gamma(t)$, which is defined to be positive/negative outside/inside the area bounded by the curve;
\item $d_i(t) := d[\bldr_i(t),t]$, signed distance from robot $i$ to the curve $\Gamma(t)$;
\item $\boldsymbol{w}^{\bot}$, vector $\boldsymbol{w}$ rotated through $+\pi/2$; positive angles are counted counterclockwise;
\item $\boldsymbol{\tau}(\bro,t)$, unit vector that is tangent to $\Gamma(t)$ at $\bro \in \Gamma(t)$ and gives the counterclockwise orientation of $\Gamma(t)$;
\item $\bn(\bro,t) = \btau(\bro,t)^{\bot}$, unit vector normal to $\Gamma(t)$;
\item $W_\tau(\bro,t) := \spr{W}{\btau(\bro,t)}, W_n(\bro,t) := \spr{W}{\bn(\bro,t)}$, tangential and normal projections of vector $W$;
\item $\st_i := \spr{\dot{\bro}_i(t)}{\btau[\bro_i(t),t]}$, tangential speed of the projection $\bro_i(t)$;
\item $\varkappa(\bro,t)$, signed curvature of the curve $\Gamma(t)$ at point $\bro \in \Gamma(t)$ at time $t$;
\item $\omega(\bro,t)$, angular velocity at which the curve $\Gamma$ rotates at point $\bro \in \Gamma(t)$ at time $t$, i.e.,
    \begin{equation}
    \label{def.omega}
    \omega(\bro,t) := \spr{\left. \frac{d \btau[\pi(\bro, t+\theta), t+\theta]}{d \theta}\right|_{\theta=0}}{\bn(\bro,t)};
    \end{equation}
    \item $\epsilon(\bro,t)$, angular acceleration of the curve $\Gamma$ at point $\bro \in \Gamma(t)$ at time $t$, i.e.,
        \begin{equation}
        \label{def.eps}
        \epsilon(\bro,t) := \left. \frac{d\omega[\pi(\bro, t+\theta), t+\theta]}{d \theta}\right|_{\theta=0}
        ;
        \end{equation}
\item $\wp(\bro,t)$, rate of change in the curvature of the curve at point $\bro \in \Gamma(t)$ at time $t$, i.e.,
$$
\wp(\bro,t) := \left. \frac{d\varkappa[\pi(\bro, t+\theta), t+\theta]}{d \theta}\right|_{\theta=0};
$$
\item $\bro_\Gamma(\theta |t,\bro)$, location of the point $\bro \in \Gamma(t)$ at time $\theta$;
\item $L_t(\bro^\prime \to \bro^{\prime\prime})$, signed length of the arc of $\Gamma(t)$ from point $\bro^\prime \in \Gamma(t)$ to
$\bro^{\prime\prime} \in \Gamma(t)$, counted counterclockwise;
    \item $\varsigma(\varrho,t)$, rate of stretch of the dynamic curve $\Gamma$ at point $\bro \in \Gamma(t)$, i.e.,
\begin{equation}
\label{def.rstretch}
\varsigma(\varrho,t):= \lim_{\theta \to t, \delta \to 0}\frac{L_\theta[\bro_\Gamma(\theta |t,\bro) \to \bro_\Gamma(\theta |t,\bro_\delta)] - \delta}{(\theta-t) \delta},
\end{equation}
where $\bro_\delta \in \Gamma(t)$ is the point of $\Gamma(t)$ at a distance of $\delta =L_t[\bro \to \bro_\delta]$ from point $\bro$;
\item $s$, natural parameter (arc length) on the curve $\Gamma(t)$, it increases when running $\Gamma(t)$ counterclockwise;
\item $f_{\bro}^\prime(\bro,t)$, derivative of the function $f(\cdot)$ defined on $\Gamma(t)$ with respect to $s$ at point $\bro \in \Gamma(t)$;
\item $\oint \ldots ds $, curvilinear integral along $\Gamma(t)$ (integral with respect to the arc length);
\item $\oms(\theta)$, infinitesimal function that has a higher order of smallness with respect to $\theta$ as $\theta$ tends to $0$;
\item $\mathbb{W}^{k,p}_{\text{loc}}(\Delta \to \br^s)$, Sobolev space of functions $g(\cdot) : \Delta \to \br^s$ that are defined and $(k-1)$ times differentiable on a (finite or infinite) interval $\Delta \subset \br$ and are such that their $(k-1)$th derivative is the antiderivative of a function from $\mathbb{L}_p$ on any compact subinterval $\Delta_{\text{com}} \subset \Delta$.
\end{itemize}
The first two results offer geometric and kinematic facts about the moving curve itself and its associates.
\begin{theorem}
\label{th.corr}
Suppose that Assumptions~{\rm \ref{ass.obs}} and {\rm \ref{ass.obs1}} are true. Then the following statements are true:
\begin{enumerate}[{\bf i)}]
\item The functions $\ddd(\cdot,\cdot)$ and $d(\cdot,\cdot)$ are continuous;
\item Suppose that for any point $(\bldr,t)$ of a set $M \subset \{ (\bldr,t) : \bldr \in \br^2, t \in \br \}$, the distance from $\bldr$ to the curve $\Gamma(t)$ is furnished by a single point of this curve. Then the function $\pi(\cdot,\cdot)$ is well-defined and continuous on $M$.
\item Suppose that the set $M$ from {\bf ii)} is open and for any its point $(\bldr,t)$, the following inequality holds
\begin{equation}
\label{non-focal}
1+ \varkappa[\pi(\bldr,t),t] d(\bldr,t) \neq 0.
\end{equation}
Then the functions $d(\cdot,\cdot)$ and $\pi(\cdot,\cdot)$ are of class $C^2$ on $M$.
\end{enumerate}
\end{theorem}
\pf
{\bf i)} The function $\mathpzc{d}(\bldr,t) = \min_{\bro \in \Gamma(t)} \|\bldr-\bro\|$ is Lipschitz continuous with respect to $\bldr$ since
$$
|\|\bldr^\prime - \bro\| -  \|\bldr^{\prime\prime} - \bro\|| \leq \|\bldr^\prime - \bldr^{\prime\prime}\| \Rightarrow |\mathpzc{d}(\bldr^\prime,t) - \mathpzc{d}(\bldr^{\prime\prime},t)| \leq \|\bldr^\prime - \bldr^{\prime\prime}\|.
$$
By invoking $\Gamma_{\text{ref}}$ and $\Phi(\cdot,\cdot)$ from Asm.~\ref{ass.obs}, we also see that
\begin{gather*}
\ddd(\bldr,t) = \min_{\bro \in \Gamma_{\text{ref}}} \|\bldr - \Phi(\bro,t)\|,
\\
|\ddd(\bldr^\prime,t^\prime)-\ddd(\bldr^\pp,t^\pp)| \leq |\ddd(\bldr^\prime,t^\prime)-\ddd(\bldr^\pp,t^\prime)| + |\ddd(\bldr^\pp,t^\prime)-\ddd(\bldr^\pp,t^\pp)|
\\
\leq \|\bldr^\prime - \bldr^\pp\| + \left|  \min_{\bro \in \Gamma_{\text{ref}}} \|\bldr^\pp - \Phi(\bro,t^\prime)\|  - \min_{\bro \in \Gamma_{\text{ref}}} \|\bldr^\pp - \Phi(\bro,t^\pp)\|\right|
\leq \|\bldr^\prime - \bldr^\pp\| + \max_{\bro \in \Gamma_{\text{ref}}} |\Phi(\bro,t^\prime) - \Phi(\bro,t^\pp)| \xrightarrow{\text{Asm.~\ref{ass.obs}}} 0
\end{gather*}
as $\bldr^\prime \to \bldr^\pp$ and $t^\prime \to t^\pp$. Thus we see that the function $\ddd(\cdot,\cdot)$ is continuous. This and Asm.~\ref{ass.obs} imply the same for  $d(\cdot,\cdot)$ since
$$
d(\bldr,t) = \begin{cases}
\ddd(\bldr,t) & \text{if} \; \bldr \; \text{lies outside} \; \Gamma(t),
\\
-\ddd(\bldr,t) & \text{if} \; \bldr \; \text{lies inside} \; \Gamma(t),
\\
\pm \ddd(\bldr,t) =0 & \text{if} \; \bldr \; \text{lies on} \; \Gamma(t).
\end{cases}
$$
\par
{\bf ii)} Since $\pi(\bldr,t)$ is the point of $\Gamma(t)$ that furnishes the distance $\ddd(\bldr,t) = \|\bldr - \pi(\bldr,t)\|$ from $\bldr$ to $\Gamma(t)$, the function $\pi(\bldr,t)$ is well-defined on $M$. Suppose that it is not continuous on $M$. Then there exists a sequence $\{(\bldr_k,t_k) \}_{k=1}^\infty \subset M$, a point $(\bldr_\ast,t_\ast) \in M$, and a number $\delta>0$ such that
\begin{equation}
\label{onf.sd}
\bldr_k \to \bldr_\ast, \quad t_k \to t_\ast \quad \text{as} \; k \to \infty \qquad \text{and} \quad \| \pi(\bldr_k,t_k) - \pi(\bldr_\ast,t_\ast)\| \geq \delta \; \forall k.
\end{equation}
Since the point $\pi(\bldr_k,t_k) \in \Gamma(t_k)$, it can be represented in the form $\pi(\bldr_k,t_k) = \Phi(\bro_k,t_k)$ with some $\bro_k \in \Gamma_{\text{ref}}$ thanks to
Asm.~\ref{ass.obs}. By passing to a subsequence, if necessary, we can ensure existence of the limit $\bro_\dagger = \lim_{k \to \infty} \bro_k \in \Gamma_{\text{ref}}$. For $\bldr_\dagger := \Phi(\bro_\dagger,t_\ast)$, we have
\begin{gather*}
\bldr_\dagger \in \Gamma(t_\ast), \quad \bldr_\dagger = \lim_{k \to \infty} \Phi(\bro_k,t_k) = \lim_{k \to \infty} \pi(\bldr_k,t_k),
\\
\|\bldr_\ast - \bldr_\dagger\|  =
\lim_{k \to \infty} \|\bldr_k - \Phi(\bro_k,t_k)\|  = \lim_{k \to \infty} \|\bldr_k - \pi(\bldr_k,t_k)\|
= \lim_{k \to \infty} \ddd(\bldr_k,t_k) \overset{\text{\bf i)}}{=} \ddd(\bldr_\ast,t_\ast).
\end{gather*}
Thus we see that the point $\bldr_\dagger \in \Gamma(t_\ast)$ furnishes the distance from $\bldr_\ast$ to $\Gamma(t_\ast)$ and so $\bldr_\dagger = \pi(\bldr_\ast,t_\ast)$.
On the other hand, letting $k \to \infty$ in the inequality from \eqref{onf.sd} yields $\|\bldr_\dagger - \pi(\bldr_\ast,t_\ast)\| \geq \delta \Rightarrow \bldr_\dagger \neq \pi(\bldr_\ast,t_\ast)$. The contradiction obtained completes the proof of {\bf ii)}.
\par
{\bf iii)} We first note that $\btau[\Phi(\bro_{\text{ref}},t),t] = \beta \Phi^\prime_{\bro_{\text{ref}}}(\bro_{\text{ref}},t)/\|\Phi^\prime_{\bro_{\text{ref}}}(\bro_{\text{ref}},t)\|$ whenever $\bro_{\text{ref}} \in \Gamma_{\text{ref}}$,
where the multiplier $\beta = \pm 1$ does not depend on $\bro_{\text{ref}},t$ and links the counterclockwise directions on $\Gamma_{\text{ref}}$ and $\Gamma(t)$, respectively.
Hence the map
$(\bro_{\text{ref}},t) \in \Gamma_{\text{ref}} \times \br \to \btau[\Phi(\bro_{\text{ref}},t),t]$ is of class $C^2$ by Asm.~\ref{ass.obs}; so is the map $(\bro_{\text{ref}},t) \in \Gamma_{\text{ref}} \times \br \to \bn[\Phi(\bro_{\text{ref}},t),t]= \big(\btau[\Phi(\bro_{\text{ref}},t),t]\big)^\bot$. Hence the map
$$
(\bro_{\text{ref}},d,t) \in \Gamma_{\text{ref}} \times \br \times \br\to Q(\bro_{\text{ref}},d,t):= \Phi(\bro_{\text{ref}},t) - d \bn[\Phi(\bro_{\text{ref}},t), t] \in \br^2
$$
is also of class $C^2$. For $(\bldr,t) \in M$, the straight-line segment with the end-points $\bldr$ and $\pi(\bldr,t)$ is normal to $\Gamma(t)$ at point $\pi(\bldr,t)$ and has only one point $\pi(\bldr,t)$ in common with $\Gamma(t)$ since otherwise the distance from $\bldr$ to $\Gamma(t)$ would be lesser than the length of this segment $\|\bldr - \pi(\bldr,t)\| $, in violation of the definition of $\pi(\bldr,t)$. It follows that
$\bldr = \pi(\bldr,t) - d(\bldr,t) \bn[\pi(\bldr,t)]$. By Asm.~\ref{ass.obs}, there exists a unique $\bro_{\text{ref}}(\bldr,t) \in \Gamma_{\text{ref}}$ such that
\begin{equation}
\label{perehod}
\pi(\bldr,t) = \Phi[\bro_{\text{ref}}(\bldr,t),t].
\end{equation}
Here the function $\bro_{\text{ref}}(\cdot,\cdot)$ is continuous on $M$ due to {\bf ii)} and Asm.~\ref{ass.obs}.
Summarizing, we infer that
\begin{equation}
\label{first.eq}
\bldr = Q[\bro_{\text{ref}}(\bldr,t),d(\bldr,t),t] \qquad \forall (\bldr,t) \in M.
\end{equation}
Meanwhile, by invoking the Frenet-Serrat equations
\begin{equation}
\label{frenet-serr}
\btau^\prime_{\bro} = \varkappa \bn, \qquad \bn^\prime_{\bro} = - \varkappa \btau
\end{equation}
and the equation $\frac{d}{d \bro_{\text{ref}}}f[\Phi(\bro_{\text{ref}},t)] = \beta \frac{d f}{d \bro}[\Phi(\bro_{\text{ref}},t)] \left\| \Phi^\prime_{\bro_{\text{ref}}}(\bro_{\text{ref}},t)\right\|$,
we see that due to \eqref{non-focal}, the partial derivatives
$$
Q^\prime_{\bro_{\text{ref}}} = \Phi^\prime_{\bro_{\text{ref}}} - d \beta \bn^\prime_{\bro} \left\| \Phi^\prime_{\bro_{\text{ref}}}\right\| \overset{\text{\eqref{frenet-serr}}}{=} \beta (1+\varkappa d) \btau \left\| \Phi^\prime_{\bro_{\text{ref}}}\right\|,
\qquad Q^\prime_d = \bn
$$
are linearly independent at any point $(\bro_{\text{ref}},d,t) = (\bro_{\text{ref}}^\star,d^\star,t^\star)$ such that $\bro_{\text{ref}}^\star = \bro_{\text{ref}}(\bldr^\star,t^\star), d^\star = d(\bldr^\star,t^\star)$ for some $(\bldr^\star,t^\star) \in M$. So by applying the implicit function theorem \cite[Thm.~3.3.1]{KrPa02} to the equation
\begin{equation}
\label{impl.fin}
Q(\bro_{\text{ref}},d,t)- \bldr = 0
\end{equation}
in the unknowns $\bro_{\text{ref}} \in \Gamma_{\text{ref}}$ and $d \in \br$, we infer the following: There exists an open neighborhood $\mathscr{N}_{\bldr^\star,t^\star} \subset M$ of $(\bldr^\star, t^\star)$ and $\delta>0$ such that whenever $(\bldr,t) \in \mathscr{N}_{\bldr^\star,t^\star}$, equation \eqref{impl.fin} has a unique solution $(\bro_{\text{ref}},d) = [\ov{\bro}_{\text{ref}}(\bldr,t), \ov{d}(\bldr,t)]$
in the set $\{\bro_{\text{ref}} \in \Gamma_{\text{ref}}, \|\bro_{\text{ref}}-\bro_{\text{ref}}^\star\| < \delta, |d - d^\star| < \delta\}$, and this solution $[\ov{\bro}_{\text{ref}}(\cdot,\cdot), \ov{d}(\cdot,\cdot)]$ is a $C^2$-smooth function of $(\bldr,t)$. By comparing \eqref{first.eq} with \eqref{impl.fin} and taking into account that $\bro_{\text{ref}}(\bldr,t) \to \bro_{\text{ref}}^\star, d(\bldr,t) \to d^\star$ as $\bldr \to \bldr^\star, t \to t^\star$ and so $\|\bro_{\text{ref}}(\bldr,t) - \bro_{\text{ref}}^\star\| < \delta, |d(\bldr,t) - d^\star| < \delta$ if $(\bldr,t)$ is close enough to $(\bldr^\star,t^\star)$, we infer that the functions
$\bro_{\text{ref}}(\cdot,\cdot)$ and $d(\cdot,\cdot)$ are identical to $\ov{\bro}_{\text{ref}}(\cdot,\cdot)$ and $\ov{d}(\cdot,\cdot)$, respectively, in some neighborhood of $(\bldr^\star,t^\star)$ and so are of class $C^2$ there. Since the point $(\bldr^\star,t^\star) \in M$ is arbitrary, they are thereby, $C^2$-smooth on $M$. It remains to invoke \eqref{perehod} and Asm.~\ref{ass.obs}. \hfill $\Box$

\begin{theorem}
Suppose that Assumptions~{\rm \ref{ass.obs}} and {\rm \ref{ass.obs1}} are true. Then
the following relations hold at any time $t$ and for any point $\bro \in \Gamma(t)$:
\begin{gather}
\label{proj}
\pi(\bro,\theta) = \bro + V_n(\bro,t) \bn(\bro,t)(\theta-t) + \oms(\theta-t),
\\
\label{dot.ln1}
\left\| \frac{\partial \pi}{\partial \bro}(\bro,t+dt)\right\| - 1 = - \varkappa(\bro,t) V_n(\bro,t) dt + \oms(dt),
\\
\label{kappa.prir}
\varkappa[\pi(\bro,t+dt), t+dt] = \varkappa(\bro,t) + (\omega^\prime_{\bro} +  \varkappa^2 V_n) dt  + \oms(dt),
\\
\label{vn.pror}
V_n[\pi(\bro,t+dt),t+dt] = V_n[\bro,t]  - (2 \omega V_\tau - A_n + \varkappa V^2_\tau) dt    + \oms(dt) ,
\\
\label{rate.str}
\varsigma(\bro,t) = \spr{\btau(\bro,t)}{V^\prime_{\bro}(\bro,t)},
\qquad
\omega(\bro,t) = \spr{\bn(\bro,t)}{V^\prime_{\bro}(\bro,t)} - \varkappa(\bro,t) V_\tau(\bro,t).
\end{gather}
\end{theorem}
\pf
{\it Proof of \eqref{proj}:}
Let $\theta \approx t$. Then $d(\bro,\theta) \approx d(\bro,t) =0$ and by Thm.~\ref{th.corr}, the projection $\pi(\bro,\theta)$ is well defined, smoothly depends on $(\bro,\theta)$, and $\bro = \pi(\bro,\theta) + a(\theta)\bn [\pi(\bro,\theta),\theta]$ for some real $a(\theta) \in \br$. Meanwhile, $\|\bro - \pi(\bro,\theta)\| \leq |a(\theta)| \leq \|\bro - \bro_\Gamma(\theta |t,\bro)\| \leq \ov{V}|\theta-t|$, where $\ov{V} := \sup_{t,\bro} \|V(\bro,t)\| < \infty$ by Asm.~\ref{ass.obs1}. By Asm.~\ref{ass.obs}, there exists $\bro_{\text{ref}}(\theta) \in \Gamma_{\text{ref}}$ such that $\pi(\bro,\theta) = \Phi[\bro_{\text{ref}}(\theta),\theta]$, and
\begin{gather*}
\pi(\bro,\theta) - \bro = \pi(\bro,\theta) - \pi(\bro,t) = \Phi[\bro_{\text{ref}}(\theta),\theta] - \Phi[\bro_{\text{ref}}(t),t]
\\
= \Phi_{\bro_{\text{ref}}}^\prime[\bro_{\text{ref}}(\theta),\theta] \frac{d\bro_{\text{ref}}(\theta)}{d \theta} (\theta - t) + V [\bro,t] (\theta-t) + \oms(\theta-t).
\end{gather*}
Since the first addend in the last expression is tangential to $\Gamma(\theta)$, whereas $\pi(\bro,\theta) - \bro$ is normal to it, we have
$$
\pi(\bro,\theta) - \bro = \spr{V(\bro,t)}{\bn[\pi(\bro,\theta),\theta]} \bn[\pi(\bro,\theta),\theta] (\theta-t) + \oms(\theta-t) \Rightarrow \text{\eqref{proj}}.
$$
\par
{\it Proof of  \eqref{rate.str}:}
By taking into account that the points $\bro \in \Gamma(t)$ of the curve are determined by the natural parameter $\bro=\bro(s)$, we see that
\begin{gather}
\label{bro_gamma}
\bro_\Gamma(t+dt |t,\bro) = \bro + V(\bro,t) dt + \oms(dt),
\\
\nonumber
L_{t+dt}[\bro_\Gamma(t+dt |t,\ov{\bro}) \to \bro_\Gamma(t+dt |t,\ov{\bro}_\delta)] = \oint_{\ov{\bro}}^{\ov{\bro}_\delta}
\left\| \frac{\partial \bro_\Gamma}{\partial \bro}(t+dt |t,\bro)  \right\|\;d s
\\
\nonumber
\overset{\text{\eqref{bro_gamma}}}{=} \oint_{\ov{\bro}}^{\ov{\bro}_\delta}
\left\| \btau(\bro,t) + V^\prime_{\bro}(\bro,t) dt + \oms(dt) \right\| \;d s
= \oint_{\ov{\bro}}^{\ov{\bro}_\delta}
\left[ 1+ \spr{\btau(\bro,t)}{V^\prime_{\bro}(\bro,t)} dt + \oms(dt)\right]d s
 \\
 \nonumber
 \overset{\text{\eqref{def.rstretch}}}{=\!=\!\Rightarrow} \text{the first equation in \eqref{rate.str}}.
\end{gather}
We put $\mathbf{Nr} W := \frac{W}{\|W\|}$ and denote by $\mathbf{Pr}_V$ the orthogonal projection onto the line spanned by $V$.
We also invoke the relation
\begin{equation}
\label{dif.norm}
\frac{d}{dt} \mathbf{Nr} W = \frac{\mathbf{Pr}_{W^\bot} \dot{W}}{\|W\|}.
\end{equation}
With these in mind, we have
\begin{gather}
 \frac{\partial V_n}{\partial \bro} =  \frac{\partial \spr{V}{\bn}}{\partial \bro} = \spr{V^\prime_{\bro}}{\bn} + \spr{V}{\bn^\prime_{\bro}} \overset{\text{\eqref{frenet-serr}}}{=} \spr{V^\prime_{\bro}}{\bn} - \varkappa V_\tau,
 \label{dif.vn}
\\
\nonumber
\btau[\pi(\bro,t+dt), t+dt] = \mathbf{Nr} \frac{\partial \pi(\bro,t+dt)}{\partial \bro}
\overset{\text{\eqref{frenet-serr},\eqref{proj}}}{=\!=\!=\!=\!=} \tau(\bro,t) + \mathbf{Pr}_{\bn} \left[ \frac{\partial V_n}{\partial \bro} \bn -\varkappa V_n \btau \right] dt + \oms(dt)
\\
\nonumber
= \tau(\bro,t)
+ \left[ \spr{V^\prime_{\bro}}{\bn} -\varkappa V_\tau \right] \bn  dt + \oms(dt)
\overset{\text{\eqref{def.omega}}}{=\!=\!\Rightarrow} \text{the second equation in \eqref{rate.str}}.
\end{gather}
{\it Proof of  \eqref{dot.ln1}} is via the following observations:
\begin{multline}
\label{pi.rho}
\pi^\prime_{\bro}(\bro,t+dt) \overset{\text{\eqref{frenet-serr},\eqref{proj}}}{=\!=\!=\!=\!=} \btau + \frac{\partial V_n}{\partial \bro} \bn dt - \varkappa V_n \btau dt + \oms(dt)
\overset{\text{\eqref{dif.vn}}}{=\!=\!=}
\btau + \underbrace{[ \spr{V^\prime_{\bro}}{\bn} - \varkappa V_\tau]}_{= \omega\,\text{by \eqref{rate.str}}} \bn dt - \varkappa V_n \btau dt + \oms(dt)
\\
= \tau + \omega \bn dt - \varkappa V_n \btau dt + \oms(dt) \Rightarrow \text{\eqref{dot.ln1}}.
\end{multline}
{\it Proof of  \eqref{kappa.prir}} is via the following observations:
\begin{gather}
\nonumber
\pi^{\prime\prime}_{\bro\bro}(\bro,t+dt) \overset{\text{\eqref{frenet-serr},\eqref{dif.vn},\eqref{pi.rho}}}{=\!=\!=\!=\!=\!=\!=\!=\!=}
\varkappa \bn + [\omega^\prime_{\bro} -\varkappa^2 V_n ]\bn dt - \big[ 2 \varkappa \omega + \varkappa^\prime_{\bro} V_n  \big] \btau dt  + \oms(dt);
\\
\nonumber
\varkappa[\pi(\bro,t+dt), t+dt] - \varkappa(\bro,t) \overset{\text{(a)}}{=} \frac{\spr{\pi^{\prime\prime}_{\bro\bro}(\bro,t+dt)}{\pi^\prime_{\bro}(\bro,t+dt)^\bot}}{\|\pi^\prime_{\bro}(\bro,t+dt)\|^3} - \varkappa
\\
\nonumber
 \overset{\text{\eqref{dot.ln1},\eqref{pi.rho}}}{=\!=\!=} \frac{\spr{\varkappa \bn + [\omega^\prime_{\bro} -\varkappa^2 V_n ]\bn dt - \big[ 2 \varkappa \omega + \varkappa^\prime_{\bro} V_n  \big] \btau dt}{\bn - \omega \btau dt - \varkappa V_n \bn dt}}{(1-\varkappa V_n dt)^3} - \varkappa+ \oms(dt) \Rightarrow \text{\eqref{kappa.prir}}.
\end{gather}
Here (a) is based on the well-known formula for the curvature of a parametric planar curve.
\par
{\it Proof of  \eqref{vn.pror}.} We start with the following remark:
\begin{gather}
\label{polza}
\pi[\bro,t+dt]-\bro_\Gamma[t+dt|t,\bro] \overset{\text{\eqref{proj}, \eqref{bro_gamma}}}{=\!=\!=\!=\!=\!=} [V_n \bn -V] dt + \oms(dt)
= - V_\tau \btau  dt + \oms(dt).
\end{gather}
We also note that the points $\bro \in \Gamma(t)$ and $\bro_\Gamma(t+dt|t,\bro)$ have a common ``seed'' $q_\ast$, i.e., for the configuration map $\Phi(\cdot)$ from Asm.~\ref{ass.obs}, $\bro = \Phi(\bldq_\ast,t), \bro_\Gamma(t+dt|t,\bro) = \Phi(\bldq_\ast,t+dt)$. It follows that
\begin{gather*}
V[ \bro_\Gamma(t+dt|\bro),t+dt] \overset{\text{\eqref{def.velacc}}}{=} \frac{\partial \Phi}{\partial t}[\bldq_\ast,t+dt] = \frac{\partial \Phi}{\partial t}[\bldq_\ast,t] + \frac{\partial^2 \Phi}{\partial t^2}[\bldq_\ast,t] dt + \oms(dt) \overset{\text{\eqref{def.velacc}}}{=} V(\bro,t) + A(\bro,t) dt + \oms(dt).
\end{gather*}
Furthermore, \eqref{def.omega} implies that $ \btau[\pi(\bro, t+dt), t+dt] - \btau(\bro,t) = \omega \bn dt + \so(dt)$
With this in mind, we see that
\begin{gather}
\nonumber
V_n[\pi(\bro,t+dt),t+dt] - V_n[\bro,t]
\\
\nonumber
= V_n[\pi(\bro,t+dt),t+dt] - V_n[\bro_\Gamma(t+dt|t,\bro),t+dt] + V_n[\bro_\Gamma(t+dt|t,\bro),t+dt]- V_n[\bro,t]
\\
\nonumber
\overset{\text{\eqref{dif.vn},\eqref{polza}}}{=\!=\!=\!=}
- \omega V_\tau dt + \spr{V + A dt}{\bn[\bro_\Gamma(t+dt|\bro),t+dt]} - \spr{V}{\bn} = - \omega V_\tau dt + A_n dt
\\
\nonumber
 + \spr{\bn[\bro_\Gamma(t+dt|\bro),t+dt]-\bn[\pi(\bro,t+dt),t+dt]}{V}
+ \spr{(\btau[\pi(\bro,t+dt),t+dt]-\btau)^\bot}{V}   + \oms(dt)
\\
\nonumber
\overset{\text{\eqref{def.omega},\eqref{dif.vn},\eqref{polza}}}{=\!=\!=\!=\!=\!=\!=\!=}
- \omega V_\tau dt + A_n dt
 + V_\tau \spr{\bn^\prime_{\bro} }{V}
+ \omega \spr{\bn^\bot}{V}   + \oms(dt)
\\
\nonumber
\overset{\text{\eqref{frenet-serr}}}{=}
- \omega V_\tau dt + A_n dt
 -\varkappa V_\tau \spr{\btau}{V}
- \omega \spr{\btau}{V}   + \oms(dt)
\Rightarrow \text{\eqref{vn.pror}}. \tag*{$\Box$}
\end{gather}
\par
In the remainder of this section, Asm.~\ref{ass.obs}---\ref{ass.focal} are supposed to be true.
Thanks to the second sentence from Asm.~\ref{ass.focal}, the projection $\pi(\bldr,t)$ is uniquely defined in the operational zone.
Our next result addresses the projection of an individual robot. From now on, we assume that in \eqref{1}, $\bldr_i(\cdot) \in \mathbb{W}^{2,1}_{\text{loc}}$ and the first equation holds almost everywhere.
\begin{theorem}
\label{lem.prrp}
Whenever robot $i$ moves in the operational zone $[\bldr_i(t),t] \in Z_{\text{\rm op}}$, the projection $\bro_i(t)$ of its location onto $\Gamma(t)$ and the distance $d_i(t)$ to this curve are of class $\mathbb{W}^{2,1}_{\text{\rm loc}}$ and the following relations hold:
\begin{gather}
\bro_i(t+dt) - \pi[\bro_i,t+dt] = \st_i \btau dt + \oms(dt),
\label{tan.vel1}
\\
\spr{\dot{\bro}_i}{\bn(\bro_i,t)} = V_n ,
\label{tan.vel}
\\
\label{tan.vel3}
\frac{d \btau[\bro_i(t),t]}{dt} = (\omega + \varkappa \st_i) \bn, \qquad \frac{d \bn[\bro_i(t),t]}{dt} = - (\omega + \varkappa \st_i) \btau,
\\
\label{dot.ln}
\dot{V}_n := \frac{d V_n[\bro_i(t),t]}{dt}= \st_i \omega   - \varkappa V_\tau^2 - 2 \omega V_\tau + A_n.
\end{gather}
\end{theorem}
\pf
Due to \eqref{zop} and \eqref{ineq.focus}, inequality \eqref{non-focal} is true everywhere in the operational zone. So by Asm.~\ref{ass.focal} and Thm.~\ref{th.corr} (where $M:= Z_{\text{op}}$), the projection $\pi(\bldr,t)$ is well-defined in $Z_{\text{op}}$, and the functions $\pi(\cdot,\cdot)$ and $d(\cdot,\cdot)$ are of class $C^2$ in $Z_{\text{op}}$. Since $\bro_i(t) = \pi[\bldr_i(t),t], d_i(t) = d[\bldr_i(t),t]$ and $\bldr_i(\cdot) \in \mathbb{W}^{2,1}_{\text{loc}}$, the functions $\bro_i(\cdot)$ and $d_i(\cdot)$
are of class $\mathbb{W}^{2,1}_{\text{loc}}$ on time intervals where robot $i$ moves in $Z_{\text{op}}$.
\par
{\it Proof of \eqref{tan.vel1}---\eqref{tan.vel3}:}
By Asm.~\ref{ass.obs}, $\bro_i(t) \in \Gamma(t) \; \forall t \Rightarrow \bro_i(t) = \Phi[\bro_{\text{ref}}(t),t]\;\forall t$
for some $\bro_{\text{ref}}(t) \in \Gamma_{\text{ref}}$. So $\dot{\bro}_i = \Phi^\prime_{\bro_{\text{ref}}} \dot{\bro}_{\text{ref}} + V$. Since $\Phi^\prime_{\bro_{\text{ref}}} \dot{\bro}_{\text{ref}}$ is tangential to $\Gamma(t)$, equation \eqref{tan.vel} is true, whereas \eqref{tan.vel1} holds since by \eqref{proj}, its l.h.s.
$
= (\dot{\bro}_i dt - V_n \bn) dt + \oms(dt) = (\dot{\bro}_i  - \spr{\dot{\bro}_i}{\bn} \bn) dt + \oms(dt) = \spr{\dot{\bro}_i}{\btau} \btau dt + \oms(dt)
$. Also,
\begin{gather*}
\btau[\bro_i(t+dt), t+dt] -\btau[\bro_i(t),t]
\\
=
\btau[\bro_i(t+dt), t+dt] -\btau[\pi(\bro_i(t),t+dt),t+dt]
+
\btau[\pi(\bro_i(t),t+dt),t+dt] -\btau[\bro_i(t),t],
\end{gather*}
which yields the first relation in \eqref{tan.vel3} due to \eqref{tan.vel1}, the Frenet-Serrat formulas \eqref{frenet-serr}, and the definition \eqref{def.omega} of the angular velocity $\omega$. Subjecting
this relation to the operation $^{\bot}$ gives the second equation from \eqref{tan.vel3}.
\par
{\it Proof of  \eqref{dot.ln}.}
We first note that
\begin{multline}
\label{polza1}
\bro_i(t+dt) -\bro_\Gamma[t+dt|t,\bro_i(t)]
\\
= \bro_i(t+dt) - \pi[\bro_i,t+dt] + \pi[\bro_i,t+dt] - \bro_\Gamma[t+dt|t,\bro_i(t)]
\overset{\text{\eqref{polza},\eqref{tan.vel1}}}{=\!=\!=\!=\!=\!=} (\st_i  - V_\tau) \btau  dt + \oms(dt).
\end{multline}
Hence we see that
\begin{gather}
\nonumber
\dot{V} dt + \oms (dt) = V[\bro_i(t+dt), t+dt] - V[\bro_i(t),t]
= V[\bro_i(t+dt), t+dt]  - V[\bro_\Gamma(t+dt|t,\bro_i(t)), t+dt]
\\
+ V[\bro_\Gamma(t+dt|t,\bro_i(t)), t+dt]
- V[\bro_i(t),t]
\\
\nonumber
\overset{\text{\eqref{def.velacc}, \eqref{polza1}}}{=\!=\!=\!=\!=} \left[ (\st_i - V_\tau) V^\prime_{\bro} + A \right] dt + \oms(dt) \Rightarrow
\dot{V} = (\st_i - V_\tau) V^\prime_{\bro} + A;
\\
\nonumber
\dot{V}_n \overset{\text{\eqref{tan.vel3}}}{=} \spr{\dot{V}}{\bn} - (\omega + \varkappa \st_i) V_\tau = \spr{(\st_i - V_\tau) V^\prime_{\bro} + A}{\bn}
 - (\omega + \varkappa \st_i) V_\tau
\\
\nonumber
\overset{\text{\eqref{rate.str}}}{=}
(\st_i - V_\tau) (\omega + \varkappa V_\tau) - (\omega + \varkappa \st_i) V_\tau + \spr{A}{\bn}
= \st_i \omega   - \varkappa V_\tau^2 - 2 \omega V_\tau + \spr{A}{\bn}
\Rightarrow \text{\eqref{dot.ln}}. \tag*{$\Box$}
\end{gather}
The main focus of the following theorem is on characterization of robot's kinematic parameters relative to the curve.
\begin{theorem}
\label{lem.supplmat}
Let $v_{i,\tau} = \spr{\bldv_i}{\btau}$ and $v_{i,n}= \spr{\bldv_i}{\bn}$ stand for the tangential and, respectively, normal components of the velocity $\bldv_i$ of robot $i$.
Similarly, let $a_{i,\tau} = \spr{\blda_i}{\btau}$ and $a_{i,n}= \spr{\blda_i}{\bn}$ stand for the tangential and, respectively, normal components of its acceleration $\blda_i$.
Whenever robot $i$ moves in the operational zone $[\bldr_i(t),t] \in Z_{\text{\rm op}}$,
the following relations hold:
\begin{gather}
\label{vel}
\bldv_i = [V_n - \dot{d}_i] \bn + [ \st_i (1+\varkappa d_i) + \omega d_i  ] \btau,
\\
\label{accel.norm}
\ddot{d}_i = \frac{\varkappa v_{i,\tau}^2+2 v_{i,\tau}  \omega - \omega^2 d_i}{1+\varkappa d_i}   - \varkappa V_\tau^2 - 2 \omega V_\tau + A_n - a_{i,n},
\\
\label{accekl.tau}
\dot{v}_{i,\tau}
=
a_{i,\btau} +  \frac{\omega+ \varkappa v_{i,\tau}}{1+\varkappa d_i} v_{i,n} =   a_{i,\btau} +  (\omega +\varkappa \st_i) v_{i,n}    ,
\\
\dot{v}_{i,n}
=
a_{i,n} -  \frac{\omega+ \varkappa v_{i,\tau}}{1+\varkappa d_i} v_{i,\tau}  =  a_{i,n} -  (\omega +\varkappa \st_i) v_{i,\tau},
\label{accekl.n}
\\
\label{dot.omega}
\dot{\omega} = \omega^\prime_{\bro} \st_i + \epsilon, \qquad \dot{\varkappa} = \varkappa^\prime_{\bro} \st_i + \wp.
\end{gather}
\end{theorem}
\pf {\it Proof of \eqref{vel}, \eqref{accel.norm}:}
Via differentiating $\bldr_i = \bro_i - d_i \bn(\bro_i,t)$, we see that
\begin{gather*}
\bldv_i = \dot{\bro}_i - \dot{d}_i \bn - d_i \dot{\bn} = \underbrace{\spr{\dot{\bro}_i}{\btau}}_{\st_i} \btau + \underbrace{\spr{\dot{\bro}_i}{\bn}}_{= V_n\,\text{by \eqref{tan.vel}}} \bn - \dot{d}_i \bn - d_i \dot{\bn}
\\
\overset{\text{\eqref{tan.vel3}}}{=} \st_i \btau + V_n \bn - \dot{d}_i \bn + (\omega + \varkappa \st_i) d_i \btau \Rightarrow \text{\eqref{vel}}.
\end{gather*}
It follows that
\begin{gather}
\label{sprime0}
\dot{d}_i = V_n - v_{i,n},
\\
\label{sprime}
\st_i = \frac{v_{i,\tau} - \omega d_i}{1+\varkappa d_i}=  v_{i,\tau} - \frac{\omega + \varkappa v_{i,\tau} }{1+\varkappa d_i} d_i ,
 \\
 \label{sprime1}
 \omega + \varkappa \st_i = \frac{\omega+ \varkappa v_{i,\tau}}{1+\varkappa d_i},
 \\
 \label{sprime2}
\st_i = v_{i,\tau} - (\omega + \varkappa \st_i)d_i,
\\
\nonumber
\ddot{d}_i \overset{\text{\eqref{tan.vel3}}}{=} \dot{V}_n - \spr{\blda_i}{n} + (\omega + \varkappa \st_i) v_{i,\tau}
\overset{\text{\eqref{dot.ln}}}{=}
\frac{\varkappa v_{i,\tau}^2+2 v_{i,\tau}  \omega - \omega^2 d_i}{1+\varkappa d_i}   - \varkappa V_\tau^2 - 2 \omega V_\tau + A_n - a_{i,n} \Rightarrow \text{\eqref{accel.norm}}.
\end{gather}
{\it Proof of \eqref{accekl.tau}, \eqref{accekl.n}} is via the following observations:
\begin{gather}
\nonumber
\dot{v}_{i,\tau} \overset{\text{\eqref{tan.vel3}}}{=} a_{i,\tau} + (\omega + \varkappa \st_i) v_{i,n}
\overset{\text{\eqref{sprime1}}}{=}
a_{i,\btau} +  \frac{\omega+ \varkappa v_{i,\tau}}{1+\varkappa d_i} v_{i,n}  \Rightarrow \text{\eqref{accekl.tau}},
\\
\nonumber
\dot{v}_{i,n} \overset{\text{\eqref{tan.vel3}}}{=} a_{i,n} -  (\omega +\varkappa \st_i) v_{i,\tau}
\overset{\text{\eqref{sprime1}}}{=}
a_{i,n} -  \frac{\omega+ \varkappa v_{i,\tau}}{1+\varkappa d_i} v_{i,\tau}
\Rightarrow \text{\eqref{accekl.n}}.
\end{gather}
{\it Proof of \eqref{dot.omega}} is via the following observations, where $\bro:= \bro_i(t)$:
\begin{gather}
\nonumber
\omega[\bro_i(t+dt), t+dt] - \omega[\bro,t] = \omega[\bro_i(t+dt), t+dt] -\omega[\pi(\bro,t+dt),t+dt] +\omega[\pi(\bro,t+dt),t+dt]-\omega[\bro,t]
\\
\nonumber
\overset{\text{\eqref{def.eps},\eqref{tan.vel1}}}{=\!=\!=\!=\!=} [\omega^\prime_{\bro}(\bro,t) \st_i(t) + \epsilon(\bro,t)] dt + \oms(dt)
\Rightarrow \text{the first formula in \eqref{dot.omega}}.
\end{gather}
The second formula in \eqref{dot.omega} is established likewise.
\hfill \epf
\par
The last theorem of the section addresses the length of $\Gamma(t)$ contained between the projections of two robots.
\begin{theorem}
Whenever robots $i$ and $j$ move in the operational zone, the following equations hold for the signed length $L_t[\bro_i(t) \to \bro_j(t)]$ of the arc of $\Gamma(t)$ from $\bro_i(t)$ to $\bro_j(t)$:
\begin{gather}
\label{lengh}
\frac{d}{dt} L_t[\bro_i(t) \to \bro_j(t)] = \st_j(t) - \st_i(t) - \oint_{\bro_i(t)}^{\bro_j(t)} \varkappa(\bro,t) V_n(\bro,t) ds,
\\
\frac{d^2}{dt^2} L_t[\bro_i(t) \to \bro_j(t)] =  a_{j,\btau} - a_{i,\btau} + \lambda_j - \lambda_i
\label{lengh.acc}
- \oint_{\bro_i}^{\bro_j} \left[ \omega^\prime_{\bro} V_n - 2 \omega \varkappa V_\tau +\varkappa A_n - \varkappa^2 V^2_\tau \right] \; ds, \quad \text{\rm where}
\\
\label{def.lambda}
\lambda_k:= \omega
V_n - \frac{  2 \dot{d}_k [\omega + \varkappa \st_k]
+ [\dot{\omega} + \dot{\varkappa} \st_k  + \varkappa a_{k,\tau} + (\st_k \varkappa + \omega) \varkappa
V_n ]d_k }{1+\varkappa d_k}.
\end{gather}
\end{theorem}
\pf
{\it Proof of  \eqref{lengh}} is via the following observations:
\begin{gather}
\nonumber
L_{t+dt}[\bro_i(t+dt) \to \bro_j(t+dt)] - L_t[\bro_i(t) \to \bro_j(t)]
\\
\nonumber
= L_{t+dt}[\bro_i(t+dt)\to \bro_j(t+dt)]
- L_{t+dt}[\bro_\Gamma(t+dt|t,\bro_i(t)) \to \bro_j(t+dt)]
\\
\nonumber
+ L_{t+dt}[\bro_\Gamma(t+dt|t,\bro_i(t)) \to \bro_j(t+dt)]
- L_{t+dt}[\bro_\Gamma(t+dt|t,\bro_i(t)) \to \bro_\Gamma(t+dt|t,\bro_j(t))]
\\
+
L_{t+dt}[\bro_\Gamma(t+dt|t,\bro_i(t)) \to \bro_\Gamma(t+dt|t,\bro_j(t))] - L_t[\bro_i(t) \to \bro_j(t)]
\\
\nonumber
\overset{\text{(a)}}{=}
L_{t+dt}[\bro_i(t+dt) \to \bro_\Gamma(t+dt|t,\bro_i(t))]
\\
\nonumber
- L_{t+dt}[\bro_j(t+dt) \to \bro_\Gamma(t+dt|t,\bro_j(t))]
+  dt \oint_{\bro_i(t)}^{\bro_j(t)} \varsigma(\bro,t) ds + \oms (dt)
\\
\nonumber
\overset{\text{\eqref{rate.str},\eqref{polza1}}}{=\!=\!=\!=\!=\!=\!=}
\{\st_j(t) - V_\tau[\bro_j(t),t]\} dt - \{\st_i(t) - V_\tau[\bro_i(t),t]\} dt
+  dt \oint_{\bro_i(t)}^{\bro_j(t)} \spr{\btau(\bro,t)}{V^\prime_{\bro}(\bro,t)} ds + \oms (dt)
\\
\nonumber
\overset{(b)}{=} \{\st_j(t) - V_\tau[\bro_j(t),t]\} dt - \{\st_i(t) - V_\tau[\bro_i(t),t]\} dt
+ \{ V_\tau[\bro_j(t),t] -  V_\tau[\bro_i(t),t] \}dt -  dt \oint_{\bro_i(t)}^{\bro_j(t)} \spr{\btau^\prime_{\bro}}{V} ds + \oms (dt)
\\
\nonumber
\overset{(c)}{=} \st_j(t) dt - \st_i(t) dt
 -  dt \oint_{\bro_i(t)}^{\bro_j(t)} \varkappa \spr{\bn}{V} ds + \oms (dt)
\nonumber
\Rightarrow \text{\eqref{lengh}}.
\end{gather}
Here (a) is based on \eqref{def.rstretch} and the equations $-L_{t^\prime}(\bro_3 \to \bro_1)=L_{t^\prime}(\bro_1 \to \bro_3) = L_{t^\prime}(\bro_1 \to \bro_2)+L_{t^\prime}(\bro_2 \to \bro_3) \; \forall \bro_1,\bro_2,\bro_3 \in \Gamma(t^\prime)$, (b) is based on integration by parts, and (c) uses the Frenet-Serrat formulas \eqref{frenet-serr}.
\par
{\it Proof of \eqref{lengh.acc}:}
We first note that
\begin{gather}
\nonumber
\frac{d}{dt}\st_i \overset{\text{\eqref{sprime}}}{=}  \dot{v}_{i,\tau}
-
\frac{(\dot{\omega} + \dot{\varkappa} v_{i,\tau} + \varkappa \dot{v}_{i,\tau})d_i + (\omega + \varkappa v_{i,\tau})\dot{d}_i}{1+\varkappa d_i}
+ \frac{(\omega + \varkappa v_{i,\tau})d_i(\dot{\varkappa} d_i + \varkappa \dot{d}_i)}{(1+\varkappa d_i)^2}
\\
\nonumber
\overset{\text{\eqref{accekl.tau}}}{=\!=\!=\!=}
a_{i,\btau} +  \frac{\omega+ \varkappa v_{i,\tau}}{1+\varkappa d_i} v_{i,n}
\\
\nonumber
-
\frac{\left[ \dot{\omega} + \dot{\varkappa} v_{i,\tau} + \varkappa \left( a_{i,\btau} +  \frac{\omega+ \varkappa v_{i,\tau}}{1+\varkappa d_i} v_{i,n}  \right) \right]d_i + (\omega + \varkappa v_{i,\tau})\dot{d}_i}{1+\varkappa d_i}
+ \frac{(\omega + \varkappa v_{i,\tau})d_i\big[\dot{\varkappa} d_i + \varkappa \dot{d}_i\big]}{(1+\varkappa d_i)^2}
\\
\nonumber
=
a_{i,\btau} +  \frac{\omega+ \varkappa v_{i,\tau}}{1+\varkappa d_i} v_{i,n}
\\
\nonumber
-
\frac{(\dot{\omega} + \dot{\varkappa} v_{i,\tau} + \varkappa a_{i,\btau}) d_i}{1+\varkappa d_i}
-
\frac{\varkappa d_i}{1+\varkappa d_i}  \frac{\omega+ \varkappa v_{i,\tau}}{1+\varkappa d_i} v_{i,n}
-
\frac{\omega + \varkappa v_{i,\tau}}{1+\varkappa d_i} \dot{d}_i
\\
\nonumber
+ \frac{\omega + \varkappa v_{i,\tau}}{1+\varkappa d_i} \frac{d_i(\dot{\varkappa} d_i + \varkappa \dot{d}_i)}{1+\varkappa d_i}
\\
\nonumber
\overset{\text{\eqref{sprime1}}}{=}
a_{i,\btau} +  (\omega + \varkappa \st_i ) v_{i,n}
\\
\nonumber
-
\frac{(\dot{\omega} + \dot{\varkappa} v_{i,\tau} + \varkappa a_{i,\btau}) d_i}{1+\varkappa d_i}
-
\frac{\varkappa d_i}{1+\varkappa d_i}   (\omega + \varkappa \st_i ) v_{i,n}
-
 (\omega + \varkappa \st_i ) \dot{d}_i
\\
\nonumber
+  (\omega + \varkappa \st_i ) \frac{d_i(\dot{\varkappa} d_i + \varkappa \dot{d}_i)}{1+\varkappa d_i}
\\
\nonumber
= a_{i,\btau} -
\frac{(\dot{\omega} + \dot{\varkappa} v_{i,\tau} + \varkappa a_{i,\btau}) d_i}{1+\varkappa d_i}
\\
\nonumber
+ (\omega + \varkappa \st_i ) \left[ v_{i,n} - \frac{\varkappa d_i}{1+\varkappa d_i} v_{i,n} -\dot{d}_i   + \frac{\varkappa  d_i}{1+\varkappa d_i} \dot{d}_i + \frac{\dot{\varkappa} d_i^2}{1+\varkappa d_i}\right]
\\
\nonumber
= a_{i,\btau} -
\frac{(\dot{\omega} + \dot{\varkappa} v_{i,\tau} + \varkappa a_{i,\btau}) d_i}{1+\varkappa d_i}
+ (\omega + \varkappa \st_i ) \left[\frac{v_{i,n}}{1+\varkappa d_i}  - \frac{\dot{d}_i}{1+\varkappa d_i}  + \frac{\dot{\varkappa} d_i^2}{1+\varkappa d_i}\right]
\\
\nonumber
\overset{\text{\eqref{sprime0}}}{=}
a_{i,\btau} -
\frac{(\dot{\omega} + \varkappa a_{i,\btau}) d_i}{1+\varkappa d_i}
-
\frac{\dot{\varkappa} v_{i,\tau} d_i}{1+\varkappa d_i}
+  \frac{\dot{\varkappa} (\omega + \varkappa \st_i ) d_i^2}{1+\varkappa d_i}
+ (\omega + \varkappa \st_i ) \left[\frac{V_n - \dot{d}_i}{1+\varkappa d_i}  - \frac{\dot{d}_i}{1+\varkappa d_i}\right]
\\
\nonumber a_{i,\btau} + \frac{(\omega + \varkappa \st_i)(
V_n - 2 \dot{d}_i)
-(\dot{\omega} + \varkappa a_{i,\tau})d_i  }{1+\varkappa d_i} + \frac{\dot{\varkappa}d_i}{1+\varkappa d_i}\underbrace{[ (\omega + \varkappa \st_i ) d_i -
 v_{i,\tau}]}_{= - \st_i \,\text{by \eqref{sprime2}}}
\\
\nonumber
=a_{i,\btau} + \frac{(\omega + \varkappa \st_i)(
V_n - 2 \dot{d}_i)
-(\dot{\omega} + \dot{\varkappa} \st_i  + \varkappa a_{i,\tau})d_i  }{1+\varkappa d_i}.
\end{gather}
Meanwhile, \eqref{lengh} implies that
\begin{equation}
\label{der.dhdhf}
\frac{d^2}{dt^2} L_t[\bro_i(t) \to \bro_j(t)] = \frac{d}{dt}\st_j - \frac{d}{dt}\st_i - \dot{\mathfrak{D}}, \quad \text{where}
\quad \mathfrak{D}(\theta) := \oint_{\bro_i(\theta)}^{\bro_j(\theta)} \varkappa(\bro,\theta) V_n(\bro,\theta) ds .
\end{equation}
To compute $\dot{\mathfrak{D}}$, we note that by \eqref{tan.vel1},
\begin{equation}
\label{fgf.dd}
\mathfrak{D}(t+dt) - \mathfrak{D}(t) =
\left[ \st_j \varkappa(\bro_j,t) V_n(\bro_j,t) - \st_i \varkappa(\bro_i,t) V_n(\bro_i,t)\right] dt + \mathfrak{B}+\oms(dt),
\end{equation}
where
\begin{gather}
\nonumber
\mathfrak{B}:=\oint_{\pi(\bro_i,t+dt)}^{\pi(\bro_j,t+dt)} \varkappa(\bro,t+dt) V_n(\bro,t+dt) ds - \oint_{\bro_i}^{\bro_j} \varkappa(\bro,t) V_n(\bro,t) ds
\\
\nonumber
\overset{\text{(a)}}{=}
\oint_{\bro_i}^{\bro_j} \varkappa[\pi(\bro,t+dt), t+dt] V_n[\pi(\bro,t+dt),t+dt] \left\| \frac{\partial \pi}{\partial \bro}(\bro,t+dt)\right\| ds - \oint_{\bro_i}^{\bro_j} \varkappa(\bro,t) V_n(\bro,t) ds
\\
\nonumber
=
\oint_{\bro_i}^{\bro_j} \left\{\varkappa[\pi(\bro,t+dt), t+dt] V_n[\pi(\bro,t+dt),t+dt] -\varkappa(\bro,t) V_n(\bro,t) \right\}ds
\\
\nonumber
+\oint_{\bro_i}^{\bro_j} \left\{ \left\| \frac{\partial \pi}{\partial \bro}(\bro,t+dt)\right\| - 1\right\} \varkappa V_n  ds + \oms(dt)
\\
\nonumber
\overset{\text{\eqref{dot.ln1}--\eqref{vn.pror}}}{=\!=\!=\!=\!=}
dt
\oint_{\bro_i}^{\bro_j} \left\{ \omega^\prime_{\bro} V_n  -  2 \varkappa \omega V_\tau + \varkappa A_n - \varkappa^2 V^2_\tau \right\}ds
 + \oms(dt).
\end{gather}
Here (a) is based on the change of the variable in the integral.
By invoking first, \eqref{fgf.dd} and then \eqref{der.dhdhf}, we infer that
\begin{gather*}
\dot{\mathfrak{D}} = \st_j \varkappa(\bro_j,t) V_n(\bro_j,t) - \st_i \varkappa(\bro_i,t) V_n(\bro_i,t) + \oint_{\bro_i}^{\bro_j} \left\{ \omega^\prime_{\bro} V_n  -  2 \varkappa \omega V_\tau + \varkappa A_n - \varkappa^2 V^2_\tau \right\}ds;
\\
\frac{d^2}{dt^2} L_t[\bro_i(t) \to \bro_j(t)] = \left[ \frac{d}{dt}\st_j - \st_j \varkappa(\bro_j,t) V_n(\bro_j,t)\right] - \left[ \frac{d}{dt}\st_i - \st_i \varkappa(\bro_i,t) V_n(\bro_i,t)\right]
\\
- \oint_{\bro_i}^{\bro_j} \left\{ \omega^\prime_{\bro} V_n  -  2 \varkappa \omega V_\tau + \varkappa A_n - \varkappa^2 V^2_\tau \right\}ds .
\end{gather*}
It remains to note that
$$
\frac{d}{dt}\st_i-  \st_i \varkappa V_n
= a_{i,\btau} + \frac{\omega
V_n  - 2 \dot{d}_i (\omega + \varkappa \st_i)
-(\dot{\omega} + \dot{\varkappa} \st_i  + \varkappa a_{i,\tau} + \st_i \varkappa^2 V_n)d_i }{1+\varkappa d_i} \overset{\text{\eqref{def.lambda}}}{=} a_{i,\btau} + \lambda_i ,
$$
and similarly $\frac{d}{dt}\st_j-  \st_j \varkappa V_n =  a_{j,\btau} + \lambda_j$, and to bring the pieces together.
\hfill \epf
\section{Three-dimensional sweep coverage of isosurfaces of time-varying environmental fields}
\label{sec.3d}
\setcounter{equation}{0}
\subsection{Motivation and Assumptions}
The interest in the developments of this section largely arises from the following scenario.
A single robot or a group of robots travel in 3D with a constant surge speed. The control input is constituted by pitching and yawing
rates. There is an unknown and time-varying scalar field described by a function $F(\boldsymbol{r},t) \in \br$  of time $t$ and spatial location $\bldr \in \br^3$. Starting from an occasional location, the robot (every robot in the team) should reach the moving and deforming {\it isosurface} $S_t(f_\star)$ where the field assumes a pre-specified value $f_\star$. Afterwards, the robot or the robotic team should move over $S_t(f_\star)$ and densely sweep the entirety of this isosurface for purposes of exposition, surveillance or processing. In the case of a group maneuver, this may include achieving and maintaining an effective self-distribution of the team
over the isosurface. This results in exhibiting and gaining control over the border of
the region with greater field values, which is commonly the major focus of interest.
Examples include tracking of oil spills or contaminant plumes \cite{ClaFie07tr}, detection and monitoring of harmful algae blooms \cite{PeDuJoPo12}, tracking zones of turbulence, contaminant clouds \cite{WTASZ05}, or high radioactivity level, exploration of sea salinity and temperature or hazardous weather conditions, to name just a few.
In such missions, a typical trouble is that information about the field is very scant and the collected data about it may quickly become obsolete due to time-variance of the field. So direct analysis of time-varying fields is highly relevant.
\par
\par
Like in \cite{MaHoSa3d14}, we employ the following general model of the robot's kinematics that ignores the roll motion:
\begin{equation}
\label{1b}
\dot{\bldr} = v \blde, \quad \dot{\blde} = \bldu, \quad \spr{\bldu}{\blde} =0.
\end{equation}
Here $\bldr$ is the robot's location, $\bldv$ is robot's velocity vector, $v>0$ is the constant surge speed, $\blde$ is the unit vector along the centerline of the robot, $\bldu \in \br^3$ is the control input,
and $\|\blde\|\equiv 1$ by the second and third equations.
Applicability of the model \eqref{1b} is discussed in \cite[Rem.~2.1]{MaHoSa3d14}; e.g., this model applies to fixed-wing aircraft, torpedo-like UUV's, and various rotorcraft.
\par
Remark~2.1 in \cite{MaHoSa3d14} discusses replacement of the ``abstract'' control $\bldu$ by the conventional pitching $q$ and yawing $r$ rates in various contexts, based on a one-to-one correspondence $\bldu \leftrightarrow (q,r)$.
That remark also shows that the model \eqref{1b} is applicable whenever the speed $v$ can be kept constant by a proper control, whereas the acceleration can be simultaneously manipulated within a disk perpendicular to the velocity $\bldv$ and centered at the origin; then $\blde:= \bldv/v$.  This holds for helicopters, submarine-like vehicles, and many other mechanical systems that move not necessarily in the surge direction \cite{CaChLe11,ReGeChFuLe12}.
\par
In this paper, special attention will be given to the situation where a part of the control task is related to a certain space direction, which is specified by a unit vector $\bldh \in \br^3$. For brevity and convenience of references, this vector is said to be {\it vertical} and the coordinate $h(\bldr)$ of point $\bldr$ in the direction of $\bldh$ is called the {\it altitude} of $\bldr$, though $\bldh$ may not be truly vertical and $h$ may not be the true altitude. Specifically, this direction takes part in the definition of the working zone $\mathcal{WZ}$, which is confined to a given range of altitudes $\mathpzc{H}_{al}=[\mathpzc{h}_-,\mathpzc{h}_+], \mathpzc{h}_- < \mathpzc{h}_+$.
Another example of a possible role of $\bldh$ is given by multi-robot scenarios where the robots should evenly distribute themselves over the altitudinal range $\mathpzc{H}_{al}$, thus forming the densest possible barrier in the vertical direction, whereas every robot sweeps the unknown and unsteady isosurface at its own altitude. This pattern integrates the sweep and barrier coverage schema, as are defined in the seminal paper \cite{gage:92}.
\par
For the sake of convenience, the definition of the working zone is completed in terms of the extreme values $f_- < f_+$ ($f_\star\in (f_-,f_+)$) taken by the field in this zone:
\begin{equation}
\label{m}
\mathcal{WZ} := \{(\bldr,t): f_- \leq F(\bldr,t) \leq f_+, \quad h(\bldr) \in \mathpzc{H}_{al} \}.
\end{equation}
\par
The following assumption is adopted everywhere in this section.
\begin{assumption}
\label{ass.smoothn}
In an open vicinity of the working zone \eqref{m}, the field $F(\cdot,\cdot)$ is twice continuously differentiable and is not singular, i.e., its spatial gradient is nonzero $\nabla F \neq 0$.
\end{assumption}
\subsection{Some geometric and kinematic formulas concerned with moving and deforming isosurfaces}
We use the following notations in this section:
\begin{itemize}
\item $\spr{\cdot}{\cdot}$, standard inner product in $\br^3$;
\item \mk $\times$, standard cross product in $\br^3$;
\item $[\vec{a},\vec{b},\vec{c}] := \spr{\vec{a}}{\vec{b} \times \vec{c}}$, scalar triple product of vectors $\vec{a}, \vec{b}, \vec{c} \in \br^3$;
\item \mk $\boldsymbol{r}(t)\in \br^3$, location of the robot at time $t$;
\item \mk $\boldsymbol{v}(t)\in \br^3$, its velocity at time $t$;
\item \mk $\boldsymbol{e}(t)\in \br^3$, the unit vector along its centerline at time $t$;
\item \mk $v$, robot's surge speed;
\item \mk $F(\bldr,t) \in \br$, unsteady environmental field in the space $\br^3$;
\item \mk $f(t) := F[\bldr(t),t]$, its value at the location of the robot;
\item \mk $\nabla F$, spatial gradient of the field;
\item \mk $\dd{F}$, its spatial Hessian;
\item \mk $\bldh$, ``vertical'' unit vector;
\item \mk $h(\bldr)$, coordinate of point $\bldr$ in the direction of $\bldh$;
\item \mk $h(t)= h[\bldr(t)]$, coordinate of the robot in the direction of $\bldh$;
\item \mk $S_t (f_\ast) =\{\bldr: F(\bldr,t)= f_\ast\}$, time-varying locus of points with the field value $f_\ast$ called the {\it isosurface};
\item \mk $S^{\text{hor}}_t(f_\star|\mathpzc{h}) := \{\bldr \in S_t(f_\star): h(\bldr) = \mathpzc{h} \}$, {\it horizontal section} (of the isosurface) at the altitude $\mathpzc{h}$;
\item \mk $N(\bldr,t) = \frac{\nabla F(\bldr,t)}{\|\nabla F (\bldr,t)\|}$, unit vector normal to the {\it as\-so\-ciated isosurface} (that passes through $\bldr$ at time $t$);
\item \mk $\alpha_h = \arcsin \spr{N}{\bldh}$, angle from $N$ to horizontal planes (i.e., those perpendicular to $\bldh$);
\item \mk $\vec{\tau} = \bldh \times N / \cos \alpha_h$, unit vector tangential to the horizontal section that passes through $\bldr$ at time $t$;
\item \mk $\mathpzc{H}_{al}=[\mathpzc{h}_-,\mathpzc{h}_+]$, operational range of the ``altitudes'';
\item \mk $\bldh_{\text{tan}} = (\bldh -N \sin \alpha_h)/\cos \alpha_h$, normalized projection of the unit vertical vector $\bldh$ onto the plane tangent to the associated isosurface;
\item $\fff{T}{T}$, second fundamental form of the associated isosurface, i.e., the quadratic form on the tangent plane whose value on any tangent unit vector $T$ is the signed curvature of the intersection of the associated isosurface with the plane spanned by $T$ and $N$ \cite[Ch.~4]{Krey91};
\item \mk $\bldr_+(\delta t| t,\bldr)$, nearest (to $\bldr$) point where the axis $\mathscr{A}_N$ drawn from $\bldr$ in the direction of $N$ intersects the time-displaced isosurface $S_{t+\delta t}[f_{\bldr,t}]$, where $f_{\bldr,t}:= F(\bldr,t)$;
    \item \mk $\zeta(\delta t| t,\bldr)$, its coordinate;
    \item \mk $\xi(\delta f| t,\bldr)$, coordinate of the nearest point of intersection between $\mathscr{A}_N$ and the space-displaced isosurface  $S_t(f_{\bldr,t}+\delta f)$;
    \item $\lambda(\bldr,t)$, front velocity of the isosurface, i.e.,
    $\displaystyle{\lim_{\delta t\to 0}}\frac{\zeta(\delta t| t,\bldr)}{\delta t }$;
\item $\alpha(\bldr,t)$, front acceleration of the isosurface:
    \begin{equation}
\label{alpha_def}
\alpha(r,t):= \lim_{\delta t\to 0}\frac{\lambda[\bldr_+(\delta t| t,\bldr), t+\delta t] - \lambda[\bldr,t]}{\delta t };
\end{equation}
\item $\vec{\omega}(\bldr,t)$, orbital angular velocity of the unit normal to the isosurface:
 \begin{equation}
 \label{n=omega}
 \vec{\omega}(r,t):= N(\bldr,t) \times \lim_{\delta t\to 0}\frac{N[\bldr_+(\delta t| t,\bldr), t+\delta t] - N[\bldr,t]}{\delta t }  ;
 \end{equation}
\item $\rho(\bldr,t)$, spatial density of the isosurfaces:
\begin{equation}
\label{spat.den}
\rho(\bldr,t):=\lim_{\delta f \to 0}\frac{\delta f}{\xi(\delta f| t,\bldr)};
\end{equation}
\item $ g_{\rho}(\bldr,t)$, proportional growth rate of this density with time:
\begin{equation}
\label{vrho.def}
g_{\rho}(\bldr,t):= \lim_{\delta t\to 0}\frac{\rho[\bldr_+(\delta t|t,\bldr), t+\delta t]- \rho(\bldr,t)}{\rho(t,\bldr) \delta t };
\end{equation}
\item $n_{\rho}(\bldr,t)$, normal proportional growth rate of the density:
\begin{equation}
\label{njrm_def}
n_{\rho} (\bldr,t):= \frac{1}{\rho(\bldr,t)}\lim_{\delta s \to 0}\frac{ \rho(\bldr+ N \delta s ,t) -   \rho(\bldr,t)}{\delta s};
\end{equation}
\item $ \grad\rho(\bldr,t)$, tangential proportional gradient of the density, i.e., the tangential vector such that for any tangential vector $T$, \begin{equation}
\label{beta_def}
\spr{\grad\rho}{T}= \frac{1}{\rho(\bldr,t)} \lim_{\delta s \to 0}\frac{\rho(r+ T \delta s,t ) -  \rho(r,t)}{\delta s} ;
\end{equation}
\item \mk $\mathscr{S}_{\bldr,t}(T) = - D_{T}N$, shape operator, where $D_TN$ is the derivative of $N$ in the direction $T$ tangential to the isosurface,
\item \mk $\proj_{\bldr,t} \boldsymbol{w}$, projection of the vector $\boldsymbol{w} \in \br^3$ onto the plane tangential to the associated isosurface at the point $(\bldr,t)$, i.e., $\proj_{\bldr,t} \boldsymbol{w} = \boldsymbol{w} - N(\bldr,t) \spr{\boldsymbol{w}}{N(\bldr,t)}$.
\end{itemize}
From the classic formula
\begin{equation}
\label{classicf}
\vec{a} \times (\vec{b} \times \vec{c}) = \spr{\vec{a}}{\vec{c}}\vec{b} - \spr{\vec{a}}{\vec{b}}\vec{c}
\end{equation}
and \eqref{n=omega}, it follows that
\begin{equation}
 \label{n=omega1}
 \frac{d}{d \theta}N[\bldr_+(\theta| t,\bldr), \theta] \big|_{\theta=t} =  \vec{\omega}(r,t) \times  N(r,t).
 \end{equation}
The density \eqref{spat.den} evaluates the number $K$ of the isosurfaces within the unit distance from the associated isosurface, where $K$ is
assessed by the span of the values assumed by the field within this distance;
\par
The first result shows that the above characteristics of the three-dimensional field are well-defined under Asm.~\ref{ass.smoothn} and explicitly relates them to derivatives of the field. This result and its proof are similar to their ``two-dimensional'' analogs, i.e., Lem.~3.1 and its proof from \cite{MSStech16}.
\begin{theorem}
\label{lem.relation}
The afore-introduced characteristics of the field are well-defined and the following relations hold in the working zone:
 \begin{gather}
 \label{speed1}
 \lambda=-\frac{F^\prime_{t}}{\|\nabla F \|}, \quad   \rho = \|\nabla F\|, \quad  \bldr_+(dt|t,\bldr) = \bldr + \lambda N dt + \so(dt),
 \\
 \label{alpha}
 g_{\rho} = \frac{\spr{\nabla F^\prime_{t}+\lambda \dd{F}N}{N}}{\|\nabla F \|}, \quad
 \vec{\omega}= N \times \frac{\nabla F^\prime_{t}+ \lambda  \dd{F}N } {\left\| \nabla F\right\|}, \quad
\alpha=- \frac{F^{\prime\prime}_{tt} + \lambda \left\langle \nabla F^\prime_t ; N \right\rangle}{\|\nabla F\|}  - \lambda  g_{\rho} ,
\\
 \label{tau}
\mathscr{S}_{\bldr,t}(T) = - \frac{\proj_{\bldr,t} [\dd{F}T]}{\|\nabla F\|}
, \quad \grad \rho = \frac{\proj_{\bldr,t}\left[F^{\prime\prime}N\right]}{\|\nabla F\|} ,
 \quad
 n_{\rho} =\frac{\spr{F^{\prime\prime}N}{N}}{\left\| \nabla F \right\|} .
\end{gather}
\end{theorem}
\pf
Given a point $(\bldr,t)$ of the working zone \eqref{m} and $\delta t, \zeta \in \br$, we denote
$$
N:=N(\bldr,t), \quad f_{\bldr,t} := F(\bldr,t), \quad \Xi(\delta t, \zeta) := F(t+\delta t, \bldr + \zeta N) - f_{\bldr,t}.
$$
Then the partial derivatives $\Xi^\prime_\zeta(0,0) = \spr{\nabla F(\bldr,t)}{N} = \|\nabla F(\bldr,t)\| \neq 0$ and $\Xi^\prime_{\delta t} (0,0) = F^\prime_t(\bldr,t)$. By the implicit function theorem \cite[Thm.~3.3.1]{KrPa02}, the equation $\Xi(\delta t, \zeta) = 0$ has a unique solution $\zeta=\zeta(\delta t)$ in a sufficiently small vicinity of $0$ for any sufficiently small $\delta t$, this solution smoothly depends on $\delta t$, and its derivative with respect to $\delta t$ at $\delta t =0$ equals $- \frac{\Xi^\prime_{\delta t} (0,0)}{\Xi^\prime_\zeta(0,0)}$. This implies that $\zeta(\delta t|t,\bldr) = \zeta(\delta t)$ for $\delta t \approx 0$, the speed $\lambda$ is well defined, and the first relation from \eqref{speed1} does hold.
\par
The same arguments show that the equation $\Upsilon (\delta f, \xi) := F(\bldr + \xi N, t) - f_{\bldr,t} - \delta f =0 $ has a unique solution $\xi=\xi(\delta f)$ in a sufficiently small vicinity of $0$ for any sufficiently small $\delta f$; this solution smoothly depends on $\delta f$, and its derivative with respect to $\delta f$ at $\delta f =0$ equals $ - \frac{\Upsilon^\prime_{\delta f} (0,0)}{\Upsilon^\prime_\xi(0,0)} = \frac{1}{\|\nabla F(\bldr,t)\|}$.
It follows that $\rho$ is well defined and the second relation from \eqref{speed1} holds.
\par
The third relation is immediate from the definitions of $\bldr_+(\delta t|t,\bldr), \zeta(\delta t| t,\bldr)$, and $\lambda$.
\par
To proceed, we introduce the shortcut $r_+(\delta t):=r_+(\delta t| t,\bldr)$ and note that due to \eqref{speed1},
\begin{gather}
\label{grd.exp}
\nabla F\left[\bldr_+(dt), t+dt \right]=
\nabla F[\bldr +\lambda N dt+\so(dt), t+dt]= \nabla F +[\nabla F^\prime_{t}+\lambda \dd{F}N]dt +\so(dt),
\\
\nonumber
\rho[\bldr_+(dt),t+dt] =
\left\| \nabla F\left[\bldr_+(dt), t+dt\right] \right\| = \left\| \nabla F+[\nabla F^\prime_{t}+ \lambda \dd{F} N]dt +\so(dt) \right\|
\\
= \left\| \nabla F\right\| + \frac{\left\langle \nabla F; \nabla F^\prime_{t}+\lambda\dd{F} N \right\rangle}{\left\| \nabla F\right\|} dt + \so(dt)
=
\rho + \left\langle N; \nabla F^\prime_{t}+ \lambda \dd{F} N \right\rangle dt + \so(dt) ,
\label{expan}
\end{gather}
which gives the first formula in \eqref{alpha}. We proceed by invoking \eqref{grd.exp}:
\begin{gather*}
N\left[\bldr_+(dt),t+dt\right] = \frac{\nabla F\left[\bldr_+(dt), t+dt\right]}{\left\| \nabla F\left[\bldr_+(dt), t+dt\right] \right\|} = N + \left[ \frac{\nabla F^\prime_{t}+ \lambda \dd{F} N}{\left\| \nabla F\right\|} - \frac{\nabla F}{\left\| \nabla F\right\|^3} \left\langle \nabla F; \nabla F^\prime_{t}+ \lambda  \dd{F}N \right\rangle  \right] dt+ \so(dt)
\\
=
N + \frac{1}{\left\| \nabla F\right\|}\left[\nabla F^\prime_{t}+ \lambda \dd{F} N - N \left\langle N; \nabla F^\prime_{t}+ \lambda  \dd{F}N \right\rangle  \right] dt+ \so(dt)
= N + \frac{\proj_{\bldr,t} \left[\nabla F^\prime_{t}+ \lambda  \dd{F}N \right]} {\left\| \nabla F\right\|} dt   + \so(dt);
\\
 \vec{\omega} \overset{\text{\eqref{n=omega}}}{=} N \times \frac{N[\bldr_+(dt), t+dt] - N}{dt }
 =
 N \times \frac{\nabla F^\prime_{t}+ \lambda \dd{F} N - N \left\langle N; \nabla F^\prime_{t}+ \lambda  \dd{F}N \right\rangle}{\left\| \nabla F\right\|}
 =
  N \times \frac{\nabla F^\prime_{t}+ \lambda \dd{F} N }{\left\| \nabla F\right\|}.
\end{gather*}
Thus we see that the second formula in \eqref{alpha} is true. Furthermore,
\begin{multline*}
\lambda[ \bldr_+(dt), t+dt] \overset{\text{\eqref{speed1}}}{=}
-\frac{F^\prime_{t}[\bldr_+(dt),t+dt]}{\|\nabla F[ \bldr_+(dt),t+dt]\|}
\\
\overset{\text{\eqref{expan}}}{=}\lambda - \frac{F^{\prime\prime}_{tt} dt + \left\langle \nabla F^\prime_t ; \bldr_+(dt)-\bldr \right\rangle}{\|\nabla F\|} + F^\prime_t \frac{\left\langle N; \nabla F^\prime_{t}+\lambda F^{\prime\prime} N \right\rangle}{\|\nabla F\|^2} dt +\so(dt)
\\
\overset{\text{\eqref{speed1}}}{=} \lambda - \frac{F^{\prime\prime}_{tt} + \lambda \left\langle \nabla F^\prime_t ; N \right\rangle}{\|\nabla F\|} dt - \lambda  \frac{\left\langle N; \nabla F^\prime_{t}+\lambda F^{\prime\prime} N \right\rangle}{\|\nabla F\|} dt +\so(dt) \overset{\text{(b)}}{=}
\lambda - \frac{F^{\prime\prime}_{tt} + \lambda \left\langle \nabla F^\prime_t ; N \right\rangle}{\|\nabla F\|} dt - \lambda  g_{\rho} dt +\so(dt),
\end{multline*}
where (b) follows from the first formula in \eqref{alpha}. The definition of $\alpha$ completes the proof of the third equation in \eqref{alpha}.
\par
Given a tangential vector $T$, the second equation in \eqref{tau} follows from the transformation
\begin{gather*}
\rho(\bldr+ T  ds ,t ) = \left\| \nabla F[\bldr + T ds,t] \right\| = \rho(\bldr,t) + \frac{\spr{F^{\prime\prime}T}{\nabla F}}{\left\| \nabla F \right\|} ds + \so(ds)
\\
= \rho(\bldr,t) + \spr{F^{\prime\prime}N}{T} ds + \so(ds) = \rho(\bldr,t) + \spr{\proj_{\bldr,t}\left[F^{\prime\prime}N\right]}{T} ds + \so(ds).
\end{gather*}
The third equation in \eqref{tau} is established likewise:
\begin{gather*}
\rho(\bldr+ N  ds ,t ) = \left\| \nabla F[\bldr + N ds,t] \right\| = \rho(\bldr,t) + \frac{\spr{F^{\prime\prime}N}{\nabla F}}{\left\| \nabla F \right\|} ds + \so(ds)
= \rho(\bldr,t) + \spr{F^{\prime\prime}N}{N} ds + \so(ds).
\end{gather*}
Finally,
\begin{gather*}
N[\bldr+T ds,t] = \frac{\nabla F [\bldr+T ds,t]}{\|\nabla F [\bldr+T ds,t]\|} = N [\bldr,t] + \left\{ \frac{\dd{F} T}{\|\nabla F\|} - \nabla F \frac{\spr{\nabla F}{\dd{F}T}}{\|\nabla F\|^3}\right\} dt + \so(ds)
\\
 = N [\bldr,t] + \frac{1}{\|\nabla F\|} \left\{ \dd{F} T - N \spr{N}{\dd{F}T}\right\} dt + \so(ds)
= N [\bldr,t] + \frac{\proj_{\bldr,t} [\dd{F}T]}{\|\nabla F\|} dt + \so(ds),
\end{gather*}
which means that the first equation in \eqref{tau} is true as well.  \epf
\par
The second result of this section displays useful relations among kinematic characteristics of the isosurfrace.
\begin{theorem}
\label{lem.bfor}
Let the vector $T$ be tangent to the isosurface. The following relations hold:
\begin{gather}
\label{lrho}
\lambda (\bldr+T ds,t) = \lambda + [\vec{\omega},T,N] ds + \so (ds),
\\
\label{lambdan}
\lambda (\bldr+N ds,t) = \lambda - g_{\rho} ds + \so (ds);
\\
\label{nnn}
N(\bldr+N ds, t)= N + \grad \rho ds + \so(ds).
\end{gather}
\end{theorem}
\pf
To prove \eqref{lrho}, we start with the following observations:
\begin{gather}
\label{step1}
\lambda(\bldr+T ds,t) \overset{\text{\eqref{speed1}}}{=} -\frac{F^\prime_{t}(\bldr+T ds,t)}{\|\nabla F(\bldr+ T ds,t)\|} = \lambda(\bldr,t) - \frac{\spr{\nabla F^\prime_{t}}{T}}{\|\nabla F\|} ds + F^\prime_t \frac{\spr{F^{\prime\prime}T}{N} }{\|\nabla F\|^2} ds + \so(ds)
\\
\label{step2}
\overset{\text{\eqref{speed1}}}{=} \lambda(\bldr,t) - \frac{\spr{\nabla F^\prime_{t}}{T}}{\|\nabla F\|} ds
- \lambda \frac{\spr{F^{\prime\prime}N}{T} }{\|\nabla F\|} ds + \so(ds)
=
\lambda(\bldr,t) - \frac{\spr{\nabla F^\prime_{t}+ \lambda F^{\prime\prime}N}{T}}{\|\nabla F\|} ds  + \so(ds)
\\
\nonumber
=
\lambda(\bldr,t) - \frac{\spr{\proj_{\bldr,t}\left(\nabla F^\prime_{t}+ \lambda F^{\prime\prime}N \right)}{T}}{\|\nabla F\|} ds  + \so(ds)
.
\end{gather}
Then we note that
\begin{gather*}
N \times \vec{\omega} \overset{\text{\eqref{alpha}}}{=} N \times \left[ N \times \frac{\nabla F^\prime_{t}+ \lambda  \dd{F}N } {\left\| \nabla F\right\|}\right]
\overset{\text{\eqref{classicf}}}{=} \spr{N}{\frac{\nabla F^\prime_{t}+ \lambda  \dd{F}N } {\left\| \nabla F\right\|}}N - \frac{\nabla F^\prime_{t}+ \lambda  \dd{F}N } {\left\| \nabla F\right\|}  = -  \proj_{\bldr,t} \frac{\nabla F^\prime_{t}+ \lambda  \dd{F}N } {\left\| \nabla F\right\|}.
\end{gather*}
Thus we see that
\begin{gather*}
\lambda(\bldr+T ds,t) = \lambda + \spr{T}{N \times \vec{\omega}} + \so(ds) = \lambda + [T,N,\vec{\omega}] + \so(ds) = \lambda + [\vec{\omega},T,N] + \so(ds) \Rightarrow \text{\eqref{lrho}}.
\end{gather*}
By retracing \eqref{step1} and \eqref{step2} with putting $N$ in place of $T$, we see that
\begin{gather*}
\lambda(\bldr+N ds,t) = \lambda(\bldr,t) - \frac{\spr{\nabla F^\prime_{t}+ \lambda F^{\prime\prime}N}{N}}{\|\nabla F\|} ds + \so(ds) \overset{\text{\eqref{alpha}}}{=}
\lambda - g_{\rho} ds + \so (ds),
\end{gather*}
which proves \eqref{lambdan}. Finally,
\begin{multline*}
N(\bldr+N ds,t) = \frac{\nabla F(\bldr+N ds,t)}{\|\nabla F(\bldr+N ds,t)\|} = N + \frac{F^\pp N}{\|\nabla F\|}ds - \nabla F \frac{\spr{F^\pp N}{\nabla F}}{\|\nabla F\|^3}ds + \so(ds)
 \\
 = N + \frac{F^\pp N - N \spr{F^\pp N}{N} }{\|\nabla F\|}ds + \so(ds)
 = N + \frac{\proj_{\bldr,t}[F^\pp N]}{\|\nabla F\|} ds + \so(ds) \overset{\text{\eqref{tau}}}{=} N + \grad \rho ds + \so(ds),
  \end{multline*}
which proves \eqref{nnn}. \hfill \epf
\par
The focus of the last theorem  is on characterization of kinematic parameters of the robot relative to field, its isosurface, and the vertical direction.
\begin{theorem}
\label{lem.veloc}
Suppose that at any point of the working zone, the unit normal $N$ to the associated isosurface is not vertical.
Whenever the robot moves in this zone, the following equations are true:
\begin{gather}
\label{ddot1}
\dot{h} = v \spr{\blde}{\bldh}, \qquad \ddot{h} = v \spr{\bldu}{\bldh}, \qquad \dot{f} = \rho [v \spr{N}{\blde}-\lambda],
\\
v \blde =
\label{full.vel}
\widehat{\lambda} N - \V, \qquad \text{\rm where}
\\
\widehat{\lambda} := \dot{f}^\rho + \lambda, \quad \dot{f}^\rho:= \dot{f}/\rho \quad \text{\rm and}
\quad
\V := \mp \vec{\tau} \frac{\V^\tau}{ \cos \alpha_h}
+ \frac{\widehat{\lambda} \sin \alpha_h-\dot{h}}{\cos \alpha_h} \hn,
\label{def.vast}
\\
\label{def.vtau}
\V^\tau:= \sqrt{v^2\cos^2\alpha_h- (\dot{h}^2+\widehat{\lambda}^2-2 \dot{h} \widehat{\lambda} \sin\alpha_h)};
\\
\label{dot.rho1}
\dot{\rho} = \dot{f} n_{\rho} - \rho \spr{\grad \rho}{\V} +\rho g_{\rho} ,
\\
\label{dot.nn1}
\dot{N} = \dot{f}^\rho \grad \rho + \mathscr{S}_{\bldr,t} \V +
\vec{\omega} \times N ,
\\
\label{dot.ll1}
\dot{\lambda} = \alpha - [\vec{\omega},\V,N] - \dot{f}^\rho g_{\rho},
\\
\ddot{f} = v \spr{N}{\bldu} -  \mathbf{II}\!\left[ \V; \V \right] + 2 [\vec{\omega},\V,N]  - \alpha
+\dot{f}^\rho \Big[\dot{f}^\rho n_{\rho}   - 2\spr{\grad \rho}{\V} + 2 g_{\rho} \Big].
\label{ddot2}
\end{gather}
\end{theorem}
\pf
The first two formulas in \eqref{ddot1} are immediate from \eqref{1b}, whereas the third is justified as follows:
$$
\dot{f}_i = F^\prime_t + \spr{\nabla F}{\dot{\bldr}}
\overset{\text{\eqref{1b}}}{=\!=} F^\prime_t + v \spr{\nabla F}{\blde}
\overset{\text{\eqref{speed1}}}{=\!=} \rho [-\lambda + v \spr{N}{\blde}].
$$
Since $\bldh$ and $N$ are not co-linear by the assumption of the theorem, $\bldh, N$, and $\vec{\tau} = \bldh \times N/\cos \alpha_h$ form a basis in $\br^3$ and so
$\blde = x \bldh + y N + z \vec{\tau}$. By finding $x,y,z$ based on the first and third equations in \eqref{ddot1} and the relations $\|\blde\| =1$, $\spr{\bldh}{N}=\sin \alpha_h$, we arrive at \eqref{full.vel} insofar as
\begin{gather*}
x = \frac{\dot{h} - \widehat{\lambda} \sin \alpha_h}{v\cos^2\alpha_h}, \qquad y = \frac{\widehat{\lambda} - \dot{h} \sin \alpha_h}{v\cos^2\alpha_h},
\qquad
z = \pm \sqrt{1- \frac{\dot{h}^2+\widehat{\lambda}^2-2 \dot{h} \widehat{\lambda} \sin\alpha_h}{v^2 \cos^2 \alpha_h}}.
\end{gather*}
\par
To prove \eqref{dot.rho1}, we observe that by \eqref{1b}, \eqref{full.vel}, \eqref{def.vast},
\begin{gather*}
\bldr(t+dt) = \bldr(t) + \lambda N dt + [\dot{f} N/\rho - \V]dt + \so(dt)
\overset{\text{\eqref{speed1}}}{=\!=}
\bldr_+[dt|t,\bldr(t)] + [ \dot{f} N/\rho -\V]dt + \so(dt).
\end{gather*}
On the other hand,
\begin{gather*}
\dot{\rho} dt +  \so(dt) = \rho[\bldr(t+dt), t+dt] - \rho[\bldr(t),t]
\\
= \rho\{\bldr(t+dt),t+dt \} - \rho\{\bldr_+[dt|t,\bldr(t)], t+dt\}
+ \rho\{\bldr_+[dt|t,\bldr(t)],t+dt\} - \rho[\bldr(t),t]
\\
\overset{\text{\eqref{vrho.def}}}{=\!=}
\rho[\bldr(t) + (\dot{f} N/\rho - \V )dt,t] - \rho[\bldr(t),t]+
\rho g_{\rho} dt + \so(dt)
\\
\overset{\text{\eqref{njrm_def},\eqref{beta_def}}}{=\!=\!=\!=}
\dot{f} n_{\rho} dt - \rho \spr{\grad \rho}{\V} dt+
\rho g_{\rho} dt + \so(dt) \Rightarrow \text{\eqref{dot.rho1}}.
\end{gather*}
Similarly,
\begin{gather*}
\dot{N}dt + \so(dt)= N(t+dt) - N(t)
= N\{\bldr_+[dt|t,\bldr(t)] + (\dot{f} N/\rho - \V )dt,t+dt \} - N\{\bldr_+[dt|t,\bldr(t)],t+dt\}
\\
+ N\{\bldr_+[dt|t,\bldr(t)],t+dt\} - N\{\bldr(t),t\} + \so(dt)
\\
\overset{\text{\eqref{n=omega1},\eqref{nnn}}}{=\!=\!=\!=}
\dot{f}/\rho \grad \rho dt + \mathscr{S}_{\bldr,t} \V dt +
\vec{\omega}\times N dt +\so(dt) \Rightarrow \text{\eqref{dot.nn1}};
\\
\dot{\lambda}dt + \so(dt)= \lambda(t+dt) - \lambda(t)
= \lambda[\bldr(t) + (\dot{f} N/\rho - \V )dt,t] - \lambda[\bldr(t),t]
\\
+ \lambda\{\bldr_+[dt|t,\bldr(t)],t+dt\} - \lambda\{\bldr(t),t\} + \so(dt)
\\
\overset{\text{\eqref{alpha_def},\eqref{lrho},\eqref{lambdan}}}{=\!=\!=\!=\!=\!=\!=}
- [\vec{\omega},\V,N] dt - \dot{f}/\rho g_{\rho} dt +
\alpha dt +\so(dt) \Rightarrow \text{\eqref{dot.ll1}}.
\end{gather*}
Thanks to \eqref{1b}, \eqref{dot.rho1}---\eqref{dot.ll1}, and the third equation from \eqref{ddot1},
\begin{gather*}
\ddot{f}/\rho = \dot{\rho}/\rho [v \spr{N}{\blde}-\lambda] + v \spr{\dot{N}}{\blde}+ v \spr{N}{\bldu}- \dot{\lambda}
\\
=  \left[ \dot{f}^\rho n_{\rho}   - \spr{\grad \rho}{\V} +g_{\rho} \right][ \spr{N}{v \blde}-\lambda]
+  \spr{\dot{f}^\rho \grad \rho + \mathscr{S}_{\bldr,t} \V + \vec{\omega} \times N}{v \blde}+ v \spr{N}{\bldu}
+ [\vec{\omega},\V,N] + \dot{f}^\rho g_{\rho} - \alpha
\\
\overset{\text{\eqref{full.vel}}}{=\!=}
\left[ \dot{f}^\rho n_{\rho}   - \spr{\grad \rho}{\V} +g_{\rho} \right][ \spr{N}{\widehat{\lambda} N - \V}-\lambda]
\\
+  \spr{\dot{f}^\rho \grad \rho + \mathscr{S}_{\bldr,t} \V + \vec{\omega} \times N}{\widehat{\lambda} N - \V}+ v \spr{N}{\bldu}
+[\vec{\omega},\V,N] + \dot{f}^\rho g_{\rho} - \alpha
\Rightarrow \text{\eqref{ddot2} } . \tag*{$\Box$}
\end{gather*}
\section{Comparison of solutions of differential equations and inequalities with discontinuous right-hand sides}
\label{sec.drhs}
\setcounter{equation}{0}
In this section, we establish technical facts related to ordinary differential equations with discontinuous right hand sides
\begin{equation}
\label{dif.rhs}
\dot{z} = f[z], \qquad z = z(t) \in \br.
\end{equation}
Specifically, we impose the following.
\begin{assumption}
\label{ass.disc}
The map $f:\br \to \br$ is locally Lipschitz continuous everywhere except for a point $\ov{z}$,
where it has one-sided limits $f(\ov{z}+) <0$ and $f(\ov{z}-) >0$. There exists $\delta>0$ such that this function is Lipschitz continuous on both $(\ov{z}-\delta,\ov{z})$ and $(\ov{z}, \ov{z}+\delta)$.
\end{assumption}
The last sentence implies existence of the one-sided limits $f(\ov{z}\pm)$.
\par
The solution of \eqref{dif.rhs} is meant in the Fillipov's sense, i.e., as the solution of the following differential inclusion
$$
\dot{z} \in F(z), \qquad \text{where} \quad F(z) :=
\begin{cases}
\{f(z)\} & \text{if} \; z \neq \ov{z},
\\
[f(\ov{z}+), f(\ov{z}-)] & \text{if} \; z = \ov{z},
\end{cases}
$$
and $\{a\}$ denotes the set with the single element $a$. Solutions of \eqref{dif.rhs} are compared with those of the following differential inequalitites
\begin{equation}
\label{diff.ineq}
\dot{z}_- \leq f[z_-], \qquad  \dot{z}_+ \geq f[z_+].
\end{equation}
They are meant as the solutions of the following differential inclusions
\begin{equation}
\label{dif.filin}
\dot{z}_\pm \in F_\pm(z), \qquad \text{where} \; F_-(z) :=
\begin{cases}
(-\infty,f(z)] & \text{if} \; z \neq \ov{z},
\\
(-\infty, f(\ov{z}-)] & \text{if} \; z = \ov{z},
\end{cases}
\quad
 F_+(z) :=
\begin{cases}
[f(z), +\infty) & \text{if} \; z \neq \ov{z},
\\
[f(\ov{z}+), + \infty) & \text{if} \; z = \ov{z} .
\end{cases}
\end{equation}
For any differential inclusion, its solution is meant as an absolutely continuous function that obeys the inclusion for almost all points $t$ from its domain of definition.
\par
The following main result of this section is well known in the case of differential equations with continuous right hand sides; see e.g., Thm.~4.1 in Chap. III \cite{Hart82}.
\begin{theorem}
\label{lem.comparis}
Suppose that Assumption~{\rm \ref{ass.disc}} is true and the absolutely continuous functions
$z_-(\cdot), z(\cdot), z_+(\cdot) : \mathscr{T}:=[\tau_-,\tau_+] \to \br$ solve the respective differential equation and inequalities from \eqref{dif.rhs} and \eqref{diff.ineq}.
Then
\begin{equation}
\label{to-be-pr}
z_-(\tau_-) \leq z(\tau_-) \leq z_+(\tau_-)  \Rightarrow z_-(t) \leq z(t) \leq z_+(t) \; \forall t \in \mathscr{T}.
\end{equation}
\end{theorem}
\par
The remainder of this section offers the proof of this theorem. Asm.{\rm \ref {ass.disc}} is supposed to be true from now on.
\begin{lemma}
\label{lem.trap}
The half-axis $(-\infty, \ov{z}]$ is a forward-trapping region for the first differential inequality in \eqref{diff.ineq}: for any solution $z_-(\cdot) : [\tau_-,\tau_+] \to \br$ of this inequality,
\begin{equation}
\label{impl.tbp}
z_-(\tau_-) \leq \ov{z} \Rightarrow z_-(t) \leq \ov{z} \quad \forall t \in [\tau_-,\tau_+].
\end{equation}
The half-axis $[\ov{z}, \infty)$ is a forward-trapping region for the second differential inequality in \eqref{diff.ineq}: for any solution $z_+(\cdot): [\tau_-,\tau_+] \to \br$ of this inequality,
$$
z_+(\tau_-) \geq \ov{z} \Rightarrow z_+(t) \geq \ov{z} \quad \forall t \in [\tau_-,\tau_+].
$$
\end{lemma}
\pf We focus on the first claim; the second one is established likewise. Suppose that \eqref{impl.tbp} fails to be true for some solution $z_-(\cdot)$. Then the open set $E:= \{t \in (\tau_-, \tau_+) : z_-(t) > \ov{z}\}$ is not empty. For its leftmost connected component $(\varsigma_-, \varsigma_+)$, we have
\begin{equation}
\label{impl.fd}
z_-(\varsigma_-) = \ov{z}, \qquad z_-(t) > \ov{z} \quad \forall t \in (\varsigma_-, \varsigma_+).
\end{equation}
Meanwhile, Asm.~\ref{ass.disc} implies that $f(z) < 0$ for all $z \in (\ov{z}, \ov{z}+\delta)$ provided that $\delta>0$ is small enough. Also $|z_-(t) - z_-(\varsigma_-)| < \delta$ for all $t \in (\varsigma_-, \varsigma)$ provided that $\varsigma \in (\varsigma_-, \varsigma_+)$ is close enough to $\varsigma_-$. Hence
\begin{gather*}
t \in (\varsigma_-, \varsigma) \Rightarrow \ov{z} < z_-(t) < \ov{z}+ \delta \Rightarrow f[z_-(t)] <0 \overset{\text{\eqref{dif.filin}}}{\Rightarrow} \dot{z}_-(t) <0
\\
\Rightarrow
z_-(t) = z_-(\varsigma_-) + \int_{\varsigma_-}^t \dot{z}_-(s)\; ds \overset{\text{(a)}}{=}  \ov{z} + \int_{\varsigma_-}^t \dot{z}_-(s)\; ds  \overset{\text{(b)}}{<} \ov{z},
\end{gather*}
Here (a) holds by the equation from \eqref{impl.fd}, whereas (b) violates the inequality from there.
The contradiction obtained completes the proof. \hfill \epf
\par
{\bf Proof of Theorem~\ref{lem.comparis}:} Let the premises from \eqref{to-be-pr} do hold. Thanks to Asm.~{\rm \ref{ass.disc}}, the equation $z=\ov{z}$ describes the sliding surface of the ODE \eqref{dif.rhs}. So one and only one of the following four scenarios occurs:
\begin{enumerate}[{\bf 1)}]
\item $z(t) < \ov{z} \; \forall t \in \mathscr{T}$,
\item $z(t) > \ov{z} \; \forall t \in \mathscr{T}$,
\item there exists $\tau \in [\tau_-,\tau_+]$ such that $z(t) < \ov{z} \; \forall t \in [\tau_-, \tau)$ and $z(t) \equiv \ov{z} \; \forall t \in [\tau,\tau_+]$,
\item there exists $\tau \in [\tau_-,\tau_+]$ such that $z(t) > \ov{z} \; \forall t \in [\tau_-, \tau)$ and $z(t) \equiv \ov{z} \; \forall t \in [\tau,\tau_+]$.
\end{enumerate}
We shall consider these cases separately.
\par
{\bf 1)} By Thm.~4.1 in Chap. III \cite{Hart82}, $z_-(t) \leq z(t)$ while both $z_-(t)$ and $z(t)$ remain in the domain $(-\infty, \ov{z})$. It follows that $z_-(t) \leq z(t)\; \forall t \in \mathscr{T}$. If $z_+(\tau_-) \geq \ov{z}$, then $z_+(t) \geq \ov{z} > z(t) \; \forall t \in \mathscr{T}$ by Lem.~\ref{lem.trap}. In the remaining case where $z_+(\tau_-) < \ov{z}$, the inequality $z_+(t) \geq z(t)$ holds until $z_+(t) <\ov{z}$ by Thm.~4.1 in Chap. III \cite{Hart82}. So if $z_+(t)$ does not arrive at $\ov{z}$, this inequality holds for all $t \in \mathscr{T}$. If $z_+(t)$ arrives at $\ov{z}$ at some time $\tau$, then $z_+(t) \geq \ov{z} > z(t)$ for $t \in [\tau,\tau_+]$ by Lem.~\ref{lem.trap}, whereas $z_+(t) \geq z(t) \; \forall t \in [\tau_-,\tau]$ by the foregoing.
\par
{\bf 2)} This case is handled likewise.
\par
{\bf 3)} By applying {\bf 1)} on the time interval $[\tau_-,\tau]$, we infer that $z_-(t) \leq z(t) \leq z_+(t) \;\forall t \in [\tau_-,\tau]$. So it remains to consider the case where $t \in [\tau,\tau_+]$. For such $t$'s, Lem.~\ref{lem.trap} guarantees that $z_-(t) \leq \ov{z}$ and $z_+(t) \geq \ov{z}$ since $z_-(\tau) \leq \ov{z}$ and $z_-(\tau) \geq \ov{z}$, respectively. It remains to invoke that $z(t) \equiv \ov{z}$ for these $t$'s.
\par
{\bf 4)} This case is handled likewise. \hfill \epf

\bibliographystyle{plain}
\bibliography{Hamidref}

 \end{document}